\documentclass{article}

\PassOptionsToPackage{numbers,compress}{natbib}

\usepackage[preprint]{neurips_2025}

\usepackage[utf8]{inputenc} 
\usepackage[T1]{fontenc}    
\usepackage{hyperref}       
\usepackage{url}            
\usepackage{booktabs}       
\usepackage{amsfonts,amsmath}       
\usepackage{nicefrac}       
\usepackage{microtype}      
\usepackage{xcolor}         
\usepackage{graphicx} 
\usepackage{enumitem}
\usepackage[capitalize]{cleveref}
\usepackage{tikz}
\usepackage{subcaption}
\usepackage{wrapfig}
\usepackage{float}
\usepackage{adjustbox} 

\title{TARDIS STRIDE: Spatio-Temporal Road Image Dataset and World Model for Autonomy}

\author{
  Héctor Carrión\textsuperscript{1,2,*}\qquad
  Yutong Bai\textsuperscript{1,3,*}\qquad
  Víctor A. Hernández Castro\textsuperscript{1,*} \\ \\
  \textbf{Kishan Panaganti}\textsuperscript{4} \qquad
  \textbf{Ayush Zenith}\textsuperscript{1} \qquad
  \textbf{Matthew Trang}\textsuperscript{1} \\ \\
  \textbf{Tony Zhang}\textsuperscript{1}   \qquad
  \textbf{Pietro Perona}\textsuperscript{4}   \qquad
 \textbf{Jitendra Malik}\textsuperscript{3}  \\\\
 \textsuperscript{1}Tera AI \qquad  
 \textsuperscript{2}UC Santa Cruz \qquad  
  \textsuperscript{3}UC Berkeley \qquad  \textsuperscript{4}California Institute of Technology
}

\begin{document}

\maketitle

\begingroup
  \renewcommand{\thefootnote}{*}
  \footnotetext{Equal contribution.}
\endgroup

\begin{abstract}
World models aim to simulate environments and enable effective agent behavior. However, modeling real-world environments presents unique challenges as they dynamically change across both space and, crucially, time. To capture these composed dynamics, we introduce a \textbf{S}patio-\textbf{T}emporal \textbf{R}oad \textbf{I}mage \textbf{D}ataset for \textbf{E}xploration (\textbf{STRIDE}) permuting 360º panoramic imagery into rich interconnected observation, state and action nodes. Leveraging this structure, we can simultaneously model the relationship between egocentric views, positional coordinates, and movement commands across both space and time. We benchmark this dataset via \textbf{TARDIS}, a transformer-based generative world model that integrates spatial and temporal dynamics through a unified autoregressive framework trained on \textbf{STRIDE}. We demonstrate robust performance across a range of agentic tasks such as controllable photorealistic image synthesis, instruction following, autonomous self-control, and state-of-the-art georeferencing. These results suggest a promising direction towards sophisticated generalist agents--capable of understanding and manipulating the spatial and temporal aspects of their material environments--with enhanced embodied reasoning capabilities. Training code, datasets, and model checkpoints are made available at \url{https://huggingface.co/datasets/Tera-AI/STRIDE}.
\end{abstract}

\section{Introduction}
\label{sec:intro}

Understanding and modeling dynamic environments is fundamental to developing intelligent systems that can effectively interact with the physical world.

As J.J. Gibson observed, ``We see in order to move; we move in order to see.'' --- a principle that highlights the intrinsic feedback loop between perception and action in natural interactive environmental exploration.

However, many autonomy-related works employ modular architectures, separating perception, mapping, prediction, and planning into distinct components. While these approaches can achieve high performance under specific tasks, such as georeferencing~\cite{svg,geoclip,pigeon}, their fragmented design fails to capture the inherently coupled information between perception and action in dynamic environments.

Recently, generative video models have shown a remarkable capability to understand real world dynamics and interactions through realistic video synthesis~\cite{sora, videopoet, cogvideo, blattmann2023align, esser2023structure}. However, these models are difficult to precisely control via their simple textual-interfaces. Furthermore, they are not explicitly encouraged to densely study physical real-world environments through free-play.

\begin{figure*}
    \centering
    \includegraphics[width=1\linewidth]{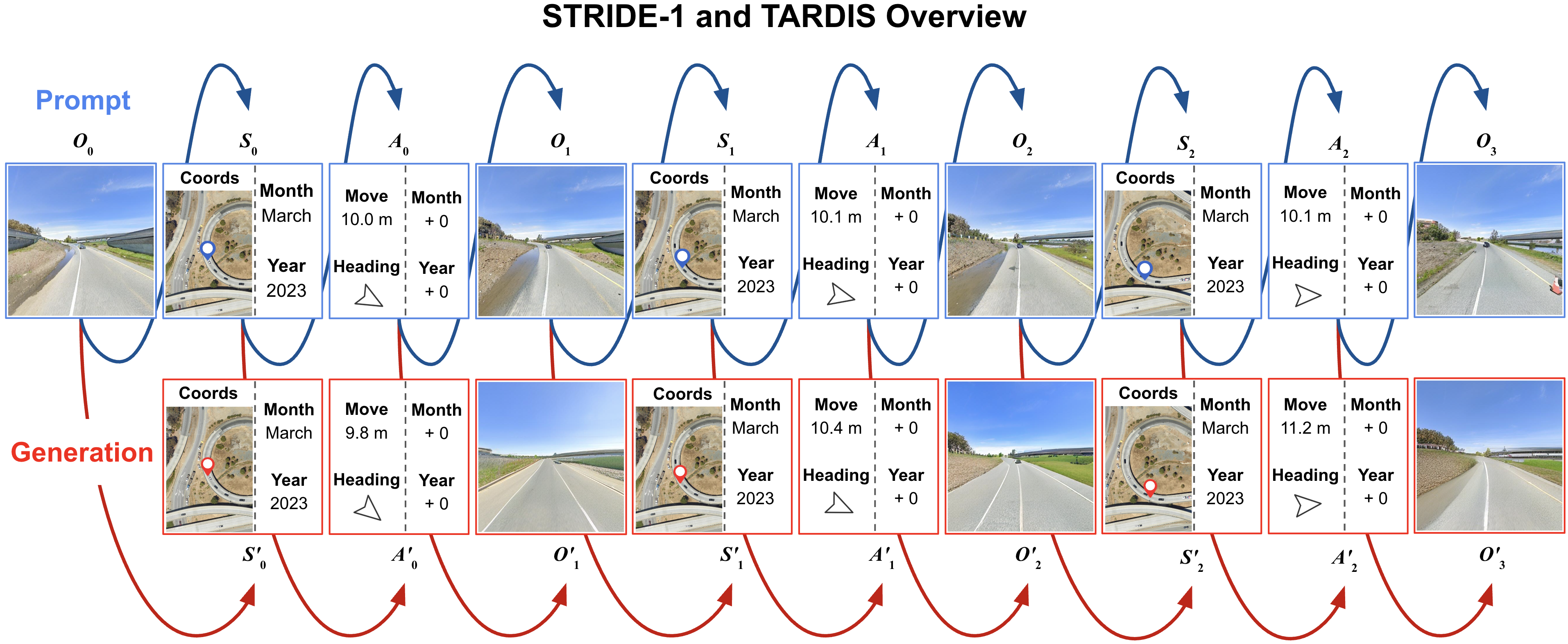}
    \vspace{-1em}
    \caption{\textbf{Data structure (STRIDE) and modeling process (TARDIS) overview.} TARDIS inputs Observation $O_0$ which conditions State $S_0$, both in space (coordinates) and time (month, year). Following this, $O_0$ and $S_0$ condition action $A_0$ spatially as a move distance in meters with heading in degrees and temporally via month, year offsets. Finally, $f_O: (O_0,S_0,A_0)\rightarrow O_1$, and the auto-regressive cycle repeats. At any point within this chain, TARDIS can be prompted with real data or asked to generate predicted data.}
    \vspace{-1em}
    \label{fig:tardi-overview}
\end{figure*}

Thus, generative world models, similarly based upon frame generation but explicitly conditioned on state-action information, have shown to be an increasingly powerful paradigm; enabling generalist agents to learn, plan, reason, and act upon their surroundings~\cite{nwm, gaia}. The primary challenge limiting these methods is the need for large amounts of coupled observation-state-action data. Because of this, recent world model work has been studied within synthetic or game environments where data can be infinitely generated based upon agent actions~\cite{gameengine, dreamer, genie}.

In addition, modeling physical environments presents a distinct challenge: unlike most video games, the real world changes over time. For example, identical spatial locations could look differently between months as seasons change. As years pass, infrastructure and housing could be newly developed, removed, renovated or otherwise updated. Understanding and thus generalizing to these natural environmental shifts could be an important aspect towards robust generative world modeling.

To address these challenges, we introduce a \textbf{S}patio-\textbf{T}emporal \textbf{R}oad \textbf{I}mage \textbf{D}ataset for \textbf{E}xploration or \textbf{STRIDE}. This novel dataset connects and projects panoramic imagery into sequential multi-modal observation-state-action events, growing the base information by over 27x without the use of traditional data augmentation. In essence, STRIDE turns real street-view panoramic observations into a navigable, interactive environment suitable for free-play across space and time.

We leverage this dataset to train TARDIS, a generative world model grounded in rich, composable spatiotemporal permutations of physical data. Our framework unifies traditionally separate challenges into a coherent sequential prediction problem, simultaneously modeling the relationship between egocentric views, positional coordinates, and movement commands across both space and time. Treating these traditionally separate problems as a single, integrated auto-regressive sequential prediction. Our main contributions are:

\begin{itemize}[nosep,leftmargin=*]
    \item \textbf{STRIDE Dataset:} A novel dataset creation method which is highly flexible, composable and efficient (generates 3.6M sequences from 131k panoramas, 27× augmentation efficiency with SSIM >0.81 temporal consistency (c.f.\,\cref{breakdown-of-data,fig:geographical-distribution-of-google-streetview})).

    \item \textbf{Controllable Spatiotemporal Generation:} TARDIS can be explicitly instructed how to move, leading to fine-grain image generation control (structural dissimilarity SSIM $<0.12$ across seasonal transitions, 41\% FID improvement over Chameleon7B (c.f.\,\cref{fig:chameleon-tardis,fig:tardis-stack})).

    \item \textbf{Advanced Georeferencing:} we include explicit latitude and longitude predictions, achieving state-of-the-art meter-level precision (60\% predictions $<10$m error vs SVG's $<$10\% at the same threshold (c.f.\,\cref{fig:50m_svg_tardis_cdf})).

    \item \textbf{Valid Self-Control:} our auto-regressive formulation allows the model to autonomously generate its own actions which we observe to be robust and consistent (77.4\% road adherence at 4m lane width, 70.5\% valid non-trivial moves (\cref{tab:model_comparison})).
    
    \item \textbf{Temporal Sensitivity:} By explicitly modeling  time as a dimension, the system can adapt and leverage temporal changes, such the ability to predict changes or estimate position despite environmental changes (linear SSIM decay R$^2$=0.94 over 5-year intervals, 18\% greater infrastructure vs seasonal distortion (c.f.\,\cref{fig:time-series-analysis-of-perplexity-with-trend-line,fig:perplexity-over-gt-temporal-action-year-and-month,fig:perplexity-over-Testing-action-move})).
    
    
    \item \textbf{Interactive Adaptation:} The flexible sequential structure enables interactive updates and refinements of predictions based on new observations (c.f.~trace interactions available on our repository.)
    
\end{itemize}

Through comprehensive evaluation, we demonstrate TARDIS's performance across multiple agentic tasks, including georeferencing, controllable photorealistic image synthesis, autonomous self-control, and a robust understanding of time.  Our results suggest that this unified approach achieves strong baseline performance in complex, time-varying environments, suggesting promising future research directions. We release our training code and model checkpoints to facilitate further study at \url{https://huggingface.co/datasets/Tera-AI/STRIDE}.

\paragraph{Paper Organization} \cref{sec:dataset-STRIDE} details our real-world dataset construction methodology, including sequential data requirements, GSV/OSM collection processes, and testing protocols. \cref{sec:tardis} introduces the TARDIS architecture, explaining its spatiotemporal modeling approach and transformer implementation. Experimental testing in \cref{sec:results} evaluates both dataset utility and model capabilities. The appendix contains extended quantitative analyses, temporal distribution visualizations, and additional generated samples.

\section{STRIDE Real-world Dataset Design}
\label{sec:dataset-STRIDE}

We develop \textbf{STRIDE} (\textbf{S}patio-\textbf{T}emporal \textbf{R}oad \textbf{I}mage \textbf{D}ataset for \textbf{E}xploration), that consists of approximately 82B tokens which were arranged into a total of 6M visual "sentences" or token sequences The sequences are generated from a relatively small set of 131k panoramic images, along with their metadata and openly available highway system data. Our approach enables a 27x increase in information, allowing for generative world model training. A complete breakdown of the dataset statistics is shown on \textbf{Table \ref{breakdown-of-data}}.

\begin{table}
  \centering
  \caption{\textbf{STRIDE Dataset.} We start with few panoramic images (131 k) but produce a sizable dataset (3.6 M) of uniquely projected visual data, all following navigational path sequences.}
  \label{breakdown-of-data}
  \begin{tabular}{lrrrr}
    \toprule
      & \textbf{Tokens} 
      & \textbf{Sequences} 
      & \textbf{Panoramas} 
      & \textbf{Projected Images} \\
    \midrule
    \textbf{Training} 
      & 80 B   
      & 5.9 M 
      & 105 k  
      & 2.9 M   \\
    \textbf{Testing}  
      & 2 B    
      & 393 k 
      & 26 k   
      & 718 k   \\
    \textbf{Total}    
      & 82 B   
      & 6.3 M 
      & 130 k  
      & 3.6 M   \\
    \bottomrule
  \end{tabular}
\end{table}

\subsection{Data for Sequential Modeling}
To allow the model to learn from the intricacies and nuances of the navigable world, we build the training data with multiple modalities that precisely describe it: images, spatial coordinates and temporal coordinates. Additionally, instructions in the form of distance and heading tokens encode movement in the physical sense while month and year offsets encode movement in a temporal sense. In summary, STRIDE represents the navigable world as a series of observations, states and actions an autonomous vehicle might experience. We carefully organize publicly available Google StreetView panoramic images along with open-source OpenStreetMap~\cite{openstreetmap2017} information to build a data graph which is consistent with the the road network.

We define ``permissible steps'' that the model is allowed to take as actions which fall within a legally drivable road segment (driveways, highways, parking lots, etc). Multi-node navigational data is created by running a DFS algorithm from each node (described in detail in the following sections), projecting views in the direction of travel and calculating states and actions along the DFS path. Additionally, we generate single-node ``look-around'' data to increase visual coverage of the surroundings and maximize exposure to the environment; this is done by projecting the aforementioned panoramic images into four 90º directions which fully captures the surroundings, this is accompanied by appropriate heading change actions. The final form of each of our navigational sequence \begin{math}V\end{math} is represented as a series of tokens that represent an array of samples at a point in space-time:
\begin{equation}
    V = \{s_{start}, N_1, ..., N_i, ..., N_n, s_{end}\}.
    \label{visual-sentence-definition-equation}
\end{equation}

Here \begin{math}s_{start}\end{math} and \begin{math}s_{end}\end{math} are the start and end sequence tokens respectively, analogous to $<bos>$ and $<eos>$ in language modeling. Each sample $N$ represents a sequence of tokens that encode an image (n=1024), spatial coordinates (n=2), temporal coordinates(n=2), and the ``next action'' (n=4) as follows with appropriate start and end tokens between to support multi-modal generation:
    \begin{align}
        N_i = \{  O_{image}, & S_{latitude}, S_{longitude}, S_{month}, S_{year}, \nonumber \\
        & A_{displacement}, A_{rotation}, A_{\Delta_{month}}, A_{\Delta_{year}} \}.
    \label{observation-definition-equation}
    \end{align}

Here, we make the distinction between ``Observation tokens'' \begin{math}O\end{math}, ``state tokens'' \begin{math}S\end{math}, and ``action tokens'' \begin{math}A\end{math}, where the first encodes visual information, the second describes the model's state in the navigable universe, while the other describes instruction in the physical and temporal axes. State tokens include the image, latitude, longitude, month and year. On the other hand, action tokens include displacement, heading, change in month and change in year. We take some freedom in notation by representing the image state tokens \begin{math}S_{image}\end{math} as \begin{math}O\end{math} interchangeably in order to distinguish these tokens as ``observations'' (defined more concretely in \cref{sec:tardis}: TARDIS).

\subsection{Data Collection}

\paragraph{Images} The images and metadata used for this experiment were obtained from open-source 360º Google StreetView (GSV) imagery. The source was chosen because its high-fidelity  and accurate metadata. Additionally, due to the nature of panoramic images, we are able to project them in a configurable number of directions, increasing the effective number of samples by a factor greater than 27. The average size of each panoramic image is around 10.0 ± 2.8 MB, and their resolutions fell into (16,384 x 8,192), (13,312 x 6,656) or (3,328 x 1,664), all with an aspect ratio of 2:1. See appendix for more details on their temporal distribution.

\begin{figure}[]
    \centering
    \includegraphics[width=0.9\linewidth]{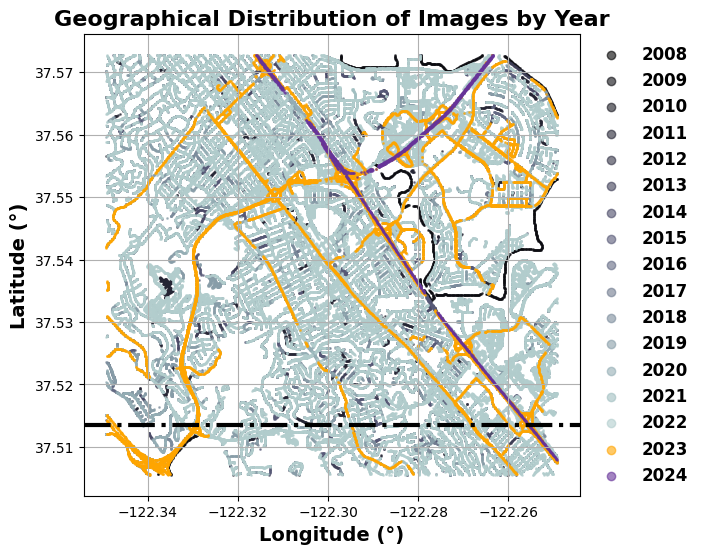}
    \caption{
    \textbf{STRIDE Geographical Distribution.} We define a 9.2km x 7.5 km area covering San Mateo City, the resulting 135k data distribution is visualized onto the map spatiotemporally. The bottom 10 percent of the map is kept out of the training set (spatiotemporal hold-out) and all observations in \textcolor{yellow}{2023} or \textcolor{violet}{2024} are also reserved for testing (temporal hold-out).}
    \label{fig:geographical-distribution-of-google-streetview}
\end{figure}

\paragraph{Multi-Node Arrangements and Road Networks} One key aspect of our dataset is that it naturally resembles a network of interconnected nodes (see \textbf{Figure \ref{fig:geographical-distribution-of-google-streetview}}), where each node in the graph is a snapshot of the surroundings in a particular point in time and space. This orderliness also happens to encode ``permissible steps'' that TARDIS would be able to take and remain in a valid spatio-temporal consistent state.

An additional benefit from arranging our GSV data into a graph is that we can permute it in arbitrary ways, which translates to a non-trivial and combinatorial amount of ``visual sequences" that we can feed into the model as training data. We limited ourselves to undirected, unweighted graphs to maximize the number of actions allowed to the model (i.e. the action ``move from \begin{math} A \rightarrow B\end{math}'' is distinctly different from ``move from \begin{math} B \rightarrow A\end{math}''), but this also means the model doesn't necessarily move along the flow of traffic. This allows the model to learn a complete state-space representation of the real world.

To permute the graph, we used a slightly modified version of iterative DFS, where we perform DFS on each node until we reach a maximum number of paths for that node or until we hit the maximum number of samples per visual sequence (both configurable parameters), whichever comes first. This was done to control the amount of compute and time spent on generating the training data.

We defined a approximately 9.2km wide by 7.5km high bounding box of San Mateo City, for which we organized 135k panoramic images. Using OpenStreetMap (OSM) data, we reconstructed the road network of the same area in order to leverage its spatial consistency for our data. We excluded all irrelevant OSM non-legal road types to only be left with valid driveways. After applying our GSV network algorithm, the result is a graph of 130k interconnected GSV nodes that we can leverage to generate permutations of permissible actions within the context of our bounded area.

\paragraph{Single-Node Sequences, Look-around Data} To complement the rich graph data, we also added so-called "look-around data", where the projections turn, completing a 360º view while on a static point in both time and space. This consisted of projecting each image inside the graph in 4 steps of 90º each. This yields \begin{math}(360^{\circ}/90^{\circ})! = 24\end{math} permutations of the same image, all with the same metadata and a static translation instruction (i.e. ``move 0 meters'') but with a nonzero look-around instruction (i.e. "turn toward 321º").

\subsection{Training \& Testing Dataset Composition}
We held-out testing data in two ways, first temporally by fully reserving data from 2023 or 2024 no matter its location (around 10.3\% of nodes). Second, we fully reserve the bottom $\sim$10\% the training area as spatiotemporal testing data (around 11.0\% of nodes). The final testing graph is comprised of around 19.8\% of the total reconstructed GSV graph's nodes. The purpose of the temporal split is to evaluate TARDIS in places where it has visited before spatially, but not during the evaluation years. The purpose of the spatiotemporal split is evaluating the model in places where it has not seen at all, either spatially or temporally. All experiments post-training are carried out within this testing data.

\section{TARDIS Model Implementation: A Spatiotemporal Modeling Process}
\label{sec:tardis}

Our approach introduces a unified framework for spatiotemporal navigation that integrates multiple traditionally separate tasks into a cohesive sequential prediction problem. At its core, the model operates through an interactive, real-time auto-regressive loop that processes observations, state coordinates, and executes navigation actions in both the spatial and temporal dimensions. 
STRIDE's graph structure enables modeling spatial (lat/lon) and temporal (month/year) states through this modeling procedure:
Observations are input as panoramic images projected toward the direction of travel, spatial state is represented as latitude and longitude coordinates, temporal state as month and year, spatial action as meter-level move distance along with true-north heading angle, and finally temporal action as month and year offsets. In this section, we formalize this modeling process below and construct our architecture inspired by it. We provide the architecture details in the appendix (\cref{sec:tardis-architecture}) as they are adopted through standard practice. 


The auto-regressive nature of our STRIDE's dataset design can be formalized through a series of inter-related sequential tasks where observation influences state which influences action and so on:

    \textbf{Spatial Localization:} Given an egocentric observation $O_n$ at step $n$, spatial-state function $f_{ss}$ determines the spatial coordinates latitude $lat$ and longitude $lon$:
    \begin{equation}\label{eq:fss}
        f_{ss}: O_n \rightarrow (lat_n, lon_n)
    \end{equation}
    This function maps raw sensory input to precise geolocation coordinates at the meter-level, serving as the foundation for subsequent positioning and navigation predictions.

    \textbf{Temporal Localization:} Given the observation $O_n$ and spatial coordinates $(lat_n, lon_n)$, temporal-state function $f_{ts}$ determines the temporal position month $m$, year $y$:
    \begin{equation}\label{eq:fts}
        f_{ts}: (O_n, lat_n, lon_n) \rightarrow (m_n, y_n)
    \end{equation}
    This enables the model to situate itself not only in space but also in time, in order to understand dynamic environmental changes. We call these spatial and temporal positions the state $S_n$.

    \textbf{Spatial Action:} Using the complete spatiotemporal context observation $O_n$ and state $S_n$, spatial action function $f_{sa}$ determines the navigational move distance in meters $d_n$ and heading in degrees $h_n$:
    \begin{equation}\label{eq:fsa}
        f_{sa}: (O_n, S_n) \rightarrow (d_n, h_n)
    \end{equation}
    This function is responsible for the commanded spatial movement which intuitively should strongly influence $O_{n+1}$ and $S_{n+1}$.

    \textbf{Temporal Action:} Here we explicitly allow action in the temporal dimension, expressed as month change $\Delta{m}_{n}$ and year change $\Delta{y}_{n}$ via temporal action function $f_{ta}$:

    \begin{equation}\label{eq:fta}
        f_{ta}: (O_n, S_n, d_n, h_n) \rightarrow (\Delta{m}_{n}, \Delta{y}_{m})
    \end{equation}
    The spatial action and temporal action space is jointly represented as $A_n$ which is the last information contained in step $n$ before moving to the next step.

    \textbf{Destination Observation:} Given information $O_n$, $S_n$, $A_n$, we generate the next observation $O_{n+1}$ at the desired spatiotemporal destination:
    \begin{equation}\label{eq:fo}
        f_{O}: (O_n, S_n, A_n) \rightarrow O_{n+1}
    \end{equation}
    This enables forward planning, as the auto-regressive cycle continues along the next function $f_{ss}$ for step $n+1$. The cycle is visualized on \textbf{Figure \ref{fig:tardi-overview}}.

This auto-regressive dynamics of TARDIS implements a Markovian process. This emerges naturally from STRIDE's graph structure \cref{fig:tardi-overview}, where node transitions depend only on immediate neighbors. To see this, for notational simplicity, let us emphasize state-action space decompositions (enabled by conditioning events denoted by $\cdot|\cdot$) mentioned through \cref{eq:fss,eq:fts,eq:fsa,eq:fta,eq:fo} as follows: At step $n$, \begin{align*}
S_n \,|\, O_n = (\underbrace{\text{lat}_n, \text{lon}_n}_{S_n^{\text{space}}}, \underbrace{m_n, y_n}_{S_n^{\text{time}}}) \,|\, O_n \quad \text{and} \quad
    A_n\,|\, \{S_n,O_n\} = (\underbrace{d_n, h_n}_{A_n^{\text{space}}}, \underbrace{\Delta m_n, \Delta y_n}_{A_n^{\text{time}}})\,|\, \{S_n,O_n\}.
\end{align*} 
In addition to state and action decompositions, the Markovian structure is further validated through observation transitions depending solely on \(S_n\) and \(A_n\), independent of history.
This Markovian structure ensures dependencies remain localized to immediate context, validated by STRIDE's \emph{OSM-grounded transitions} where:
\[
P(S_{t+1}|S_t,A_t) > 0 \iff \exists\ \text{valid road path}\ S_t \rightarrow S_{t+1}
\]
as defined through OpenStreetMap network constraints \cite{openstreetmap2017}.

\section{Experiments with TARDIS on STRIDE Dataset}
\label{sec:results}

We design experiments that probe the quality of each component of TARDIS individually and in-tandem. We note that it is possible to compose prompts with variable amounts of ground-truth data which allows the model to perform deep or shallow in-context learning.

We measure all metrics and visualize all results using purely testing data which has been kept completely outside the models training regime spatiotemporally (bottom 10 percent of the training area).

\subsection{Controllable Image Generation}

Generating photorealistic images in a physically controllable way is an under-explored task. Existing generative models either output a single image given a text description or attempt to understand commanded change via analogy prompting. TARDIS is grounded on real-world data, states and instructions, leading to controllable image generation.

\begin{figure}
    \centering
    \includegraphics[width=\linewidth]{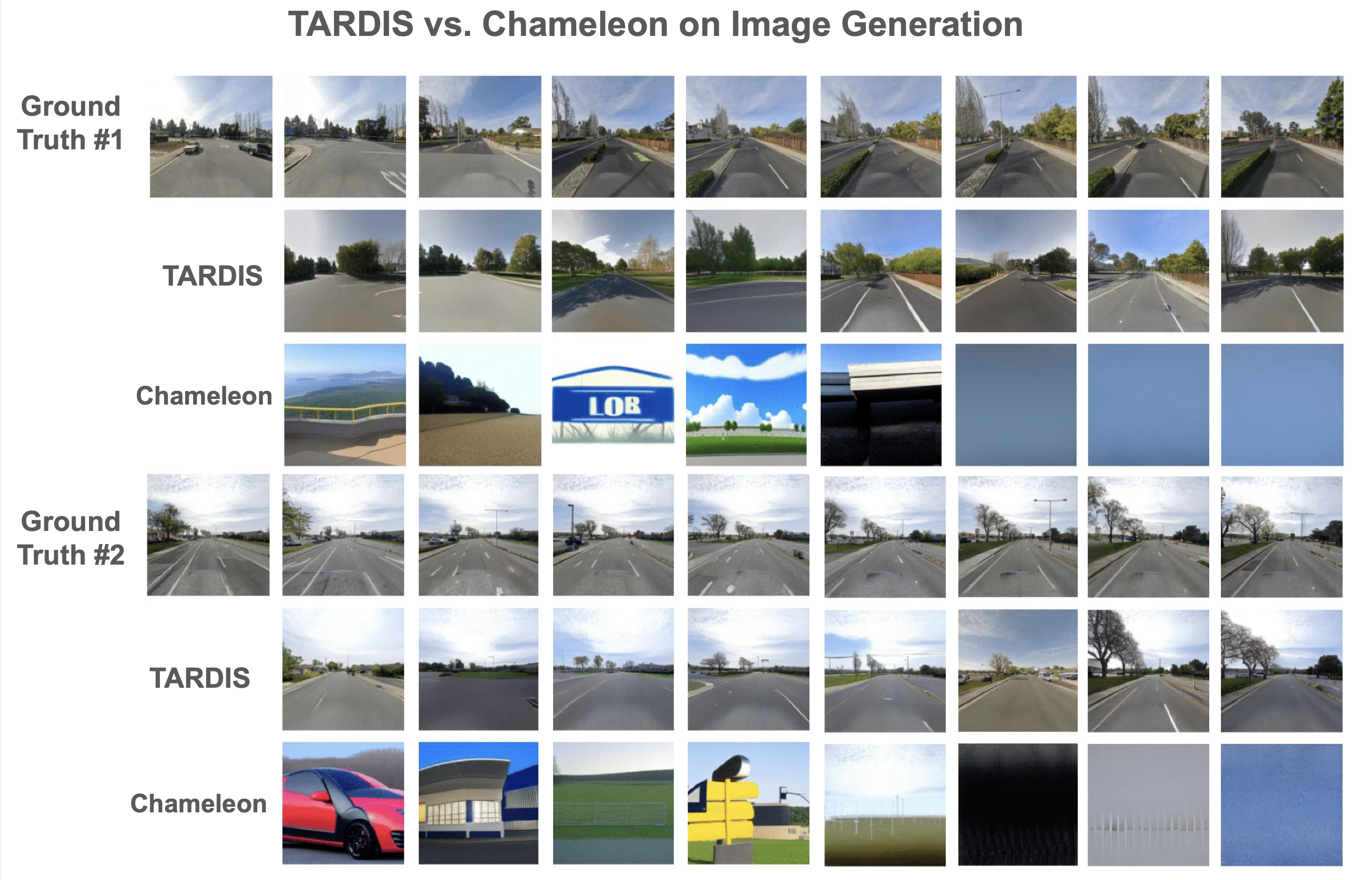}
    \caption{\textbf{TARDIS vs. Chameleon Image Generation.} TARDIS appears to more closely follow the image pattern and instructions. In addition, Chameleon smaller context length (4k vs 16k) impacts performance after 5 frames.}
    \label{fig:chameleon-tardis}
\end{figure}

We prompt TARDIS with spatio-temporally held-out testing ground-truth states and instructions in order to compare newly generated images against expected real-image results. We qualitatively compare against the recent Chameleon-7B \cite{chameleon}, another naively multi-modal architecture trained to follow textual instructions. Chameleon uses a similar (although bigger) transformer architecture as well as the same image tokenizer. To identify if STRIDE has enabled the model to understand physical space to a higher degree than general-purpose models, we prompt Chameleon with the textual instruction equivalent of our state-action paradigm along with the same image observations. \textbf{Figure \ref{fig:chameleon-tardis}} shows the difficulty in control and consistency issues of Chameleon in comparison with TARDIS. Further qualitative results can be found on \textbf{Figure \ref{fig:tardis-stack}}.

We have further visualizations in the \cref{sec:appen-additional-results} for space constraints. We measure quality quantitatively via the perplexity metric. Note that visual metrics have often been found to not necessarily be representative of image quality. However, we measure perplexity as a function of spatial and temporal commands. We find that along the breadth of temporal coverage in the testing data the perplexity trend remains mostly stable \cref{fig:time-series-analysis-of-perplexity-with-trend-line}. Further, while movement actions does have a large spread of error, the overall trend is again mostly consistent with a drop around 10m, this is likely because the majority of spatial distances in our data falls near the 10m range \cref{fig:perplexity-over-Testing-action-move}. Finally, we observe that for the majority of temporal actions both in month and year perplexity remains largely consistent with spokes at the edges (-15, +13) of the year temporal actions present in the testing data \cref{fig:perplexity-over-gt-temporal-action-year-and-month}. This, along with the visual results suggests the model is reliably following the generative instructions given to it during evaluation.

\subsection{Georeferencing} 


    
Georeferencing is the task of returning the geo location of a given input image. It is common to return the global latitude and longitude coordinates when the task is absolute positioning. TARDIS is able to output coordinate tokens which correspond to global cooridnates within its 9.2x7.5 km training and evaluation area.

We prompt TARDIS with 1...n testing nodes and measure its real-world meter error compared to the real coordinate. Since TARDIS is able to input spatial commands which relate to the following spatial location, we dynamically allow self-masking of tokens which fall outside the expected radius defined by the move command. This allows the model reliably predict coordinate tokens physically outside the spatial distribution it was trained on, in other words, predict tokens within the unseen testing area.

We compare against a leading method \cite{svg} which leverages a contrastive method for ground-image to aerial map similarity measurements. Similar to TARDIS this method was not trained on the testing area for which we measure performance and TARDIS does not input the aerial images the baseline method requires. We now describe the baseline method:
\textbf{Statewide Visual Geolocalization (SVG)}~\cite{svg} is a model which is able to predict the location of a street-view photo by matching it on a database of aerial reference imagery. The model is trained on mapping data between street and aerial data on two US states, and four German states, and uses a ConvNeXt backbone alongside attention-based pooling to generate an embedding vector for the street and aerial regions. Cosine similiarity between the query point and a reference database location is then used to geolocate the query image. 

\begin{figure}[t]
    \centering
    \includegraphics[width=\linewidth]{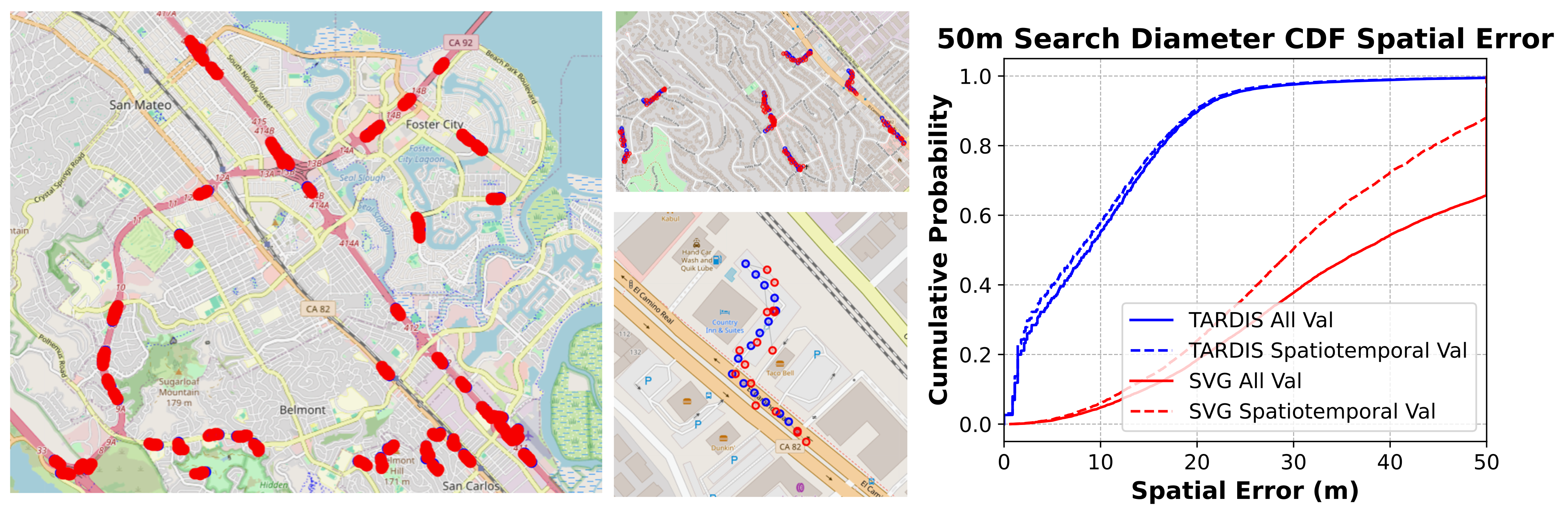}
    \caption{
    \textbf{(Left) TARDIS Georeferencing.} \textcolor{blue}{Blue} signifies ground-truth locations, \textcolor{red}{Red} signifies TARDIS coordinate predictions.
    \textbf{(Right) CDF of TARDIS vs SVG with 50 Meter Search Space.} Describes the fraction of predictions within some distance threshold. In particular, TARDIS model has 60\% of predictions within 10m, 90\% within 20m. }
    \label{fig:50m_svg_tardis_cdf}
\vspace{-2em}
\end{figure}

In \textbf{Figure \ref{fig:50m_svg_tardis_cdf}}, we demonstrate the results of a CDF on the geolocation error of the two methods, on both the entire temporally held-out dataset and the spatiotemporally held-out area data. Furthermore, we restrict SVG's ``search space'' to that of only 50m around the true location of the camera, this is to match TARDIS's maximum move action of 50 meters. According to the CDF plot in \textbf{Figure \ref{fig:50m_svg_tardis_cdf}} TARDIS achieves 60\% of predictions within 10m while SVG does not reach 10\% at the same metric. 
Moreover, in \cref{sec:appen-additional-results}, we also show the fraction of prediction errors in \cref{fig:svg_tardis_per_year} for both models within full and 50 meters search spaces. We observe a large performance advantage over SVG.

\vspace{-1em}

\subsection{Self-Control Quality}
We measure self-control quality by allowing the model to generate its own distance and heading instructions, given real testing data on fully held-out areas spatiotemporally.

The intention is to observe whether or not, for an unvisited region, the model is able to understand its surroundings given real visual observation inputs. We measure this quantitatively by calculating the percent of actions which produce a following state within some distance threshold of the nearest road center line.

Given an average lane length in the USA of 12ft (3.7m), TARDIS generated actions fall within the road 77.4\% of the time on unseen testing environments. After filtering out trivial static actions (move = 0), the valid action percentage remains above 70\%. 
To avoid inadvertently picking an optimal lane length that would maximize the model's success rate, we calculated results for lane-lengths in increments of 1m, up to 10m. We displayed all results on \textbf{\cref{tab:model_comparison}}. To ensure the model is not taking very small steps in order to stay within bounds, we observe that 60\% of the time TARDIS generates steps of 5m or greater (see~\cref{fig:cdf-of-spatial-action} in \cref{sec:appen-additional-results}), which is wider than many residential roads.\begin{wrapfigure}{r}{20em}
\vspace{-1em}
\centering
    \begin{tabular}{ccc}
\toprule
Lane Length &All Actions &Nonzero Move\\
 (m) & Valid\% &Valid\%\\
\midrule
1.0 & 55.0 & 42.4 \\
2.0 & 66.8 & 56.3 \\
3.0 & 72.8 & 64.2 \\
4.0 & 77.4 & 70.5 \\
5.0 & 81.4 & 75.8 \\
6.0 & 85.0 & 80.6 \\
7.0 & 88.0 & 84.6 \\
8.0 & 90.6 & 88.0 \\
10.0 & 94.9 & 93.8 \\
\bottomrule 
\end{tabular} 
\caption{\textbf{Self-Control Quality for Different Lane Widths.} } \label{tab:model_comparison}
\vspace{-4em}
\end{wrapfigure}.

\begin{figure*}[t]
    \centering
\includegraphics[width=\linewidth]{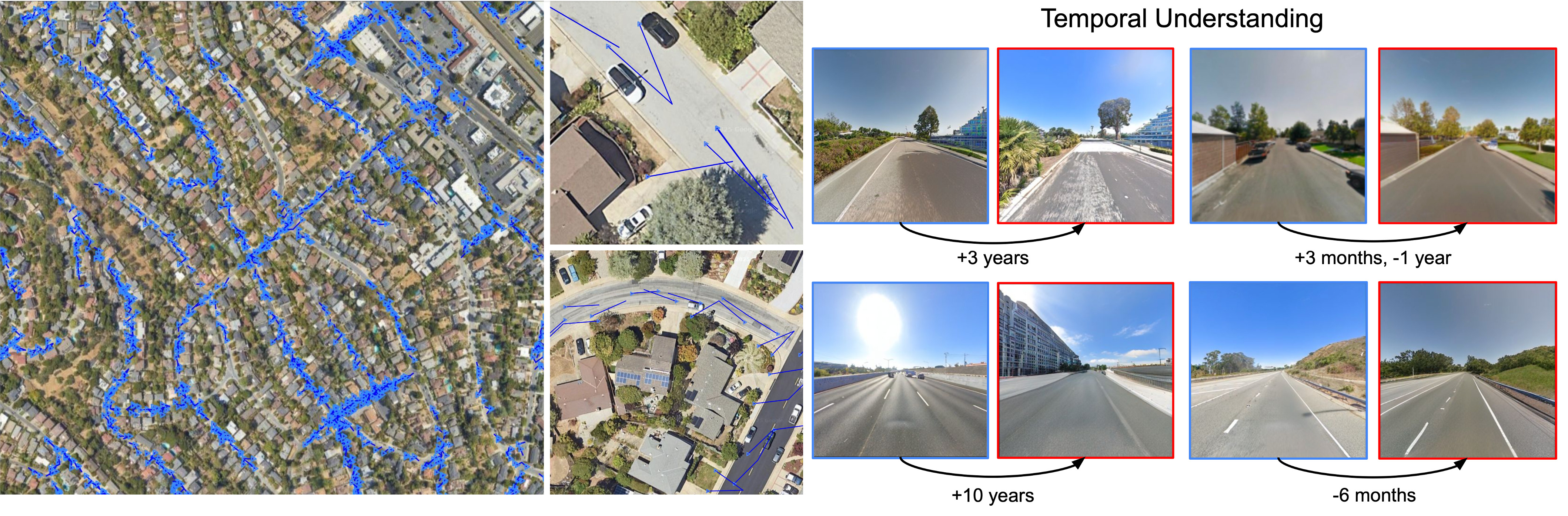}
\caption{\textbf{(Left) Qualitative Analysis of Self-Control.} Arrowheads represent heading action, line length represents distance action generated by TARDIS. We observe the majority of actions keep within the lane or onto driveways.
\textbf{(Right) Temporal Reasoning.} \textcolor{blue}{Blue} represents a prompt image, \textcolor{red}{red} represents the generated output image given a manual temporal instruction. Top right example moves from Autum (September 2014) to Winter (December 2013). Bottom right action moves 6 months in order to change seasons.
    }     \label{fig:model-following-road-network-examples}
\vspace{-2em}
\end{figure*}

For qualitative analysis we plot heading (arrowhead) and distance (line length) on a satellite map of the spatiotemporally held-out area, shown in \textbf{Figure \ref{fig:model-following-road-network-examples}}. Our observations show the majority of the actions fall within the road network, with lane and curve following. We note sometimes the generated action take the next state into a driveway which is not a behavior present in the GSV or STRIDE dataset, but could be an emergent property.


\subsection{Sense of time}

In order to measure the models understanding of temporal dynamics we manually prompt a static spatial move along with large temporal change on spatiotemporally held-out testing data. The objective is to advance time enough to observe seasonal change, observe if transient objects have disappeared, or other large changes while simultaneously the generated image remains largely consistent. We qualitatively provide some examples of this property on \textbf{Figure \ref{fig:model-following-road-network-examples}}, advancing from Fall to Winter does appear to affect the foliage (right columns) while larger time skips impact the infrastructure with degrading roads or new developments appearing (left columns).


\section{Conclusion}
\label{sec:conclusion}
We introduce emergent properties, a robust model and a novel dataset that could serve the research community towards future real-world-models that can handle temporally dynamic settings in unseen environments. As generalist agents transition away from synthetic environments and onto the real world tools like TARDIS and STRIDE could enable novel tasks completion.

{
  \small
  \bibliographystyle{unsrtnat}   
  \bibliography{example_paper,neurips_extra}    

\begin{thebibliography}{56}
\providecommand{\natexlab}[1]{#1}
\providecommand{\url}[1]{\texttt{#1}}
\expandafter\ifx\csname urlstyle\endcsname\relax
  \providecommand{\doi}[1]{doi: #1}\else
  \providecommand{\doi}{doi: \begingroup \urlstyle{rm}\Url}\fi

\bibitem[Fervers et~al.(2024)Fervers, Bullinger, Bodensteiner, Arens, and Stiefelhagen]{svg}
Florian Fervers, Sebastian Bullinger, Christoph Bodensteiner, Michael Arens, and Rainer Stiefelhagen.
\newblock Statewide visual geolocalization in the wild.
\newblock In \emph{ECCV}, 2024.

\bibitem[Vivanco~Cepeda et~al.(2024)Vivanco~Cepeda, Nayak, and Shah]{geoclip}
Vicente Vivanco~Cepeda, Gaurav~Kumar Nayak, and Mubarak Shah.
\newblock Geoclip: Clip-inspired alignment between locations and images for effective worldwide geo-localization.
\newblock \emph{Advances in Neural Information Processing Systems}, 36, 2024.

\bibitem[Haas et~al.(2024)Haas, Skreta, Alberti, and Finn]{pigeon}
Lukas Haas, Michal Skreta, Silas Alberti, and Chelsea Finn.
\newblock Pigeon: Predicting image geolocations.
\newblock In \emph{Proceedings of the IEEE/CVF Conference on Computer Vision and Pattern Recognition}, pages 12893--12902, 2024.

\bibitem[Brooks et~al.(2024)Brooks, Peebles, Holmes, DePue, Guo, Jing, Schnurr, Taylor, Luhman, Luhman, et~al.]{sora}
Tim Brooks, Bill Peebles, Connor Holmes, Will DePue, Yufei Guo, Li~Jing, David Schnurr, Joe Taylor, Troy Luhman, Eric Luhman, et~al.
\newblock Video generation models as world simulators, 2024.

\bibitem[Kondratyuk et~al.(2023)Kondratyuk, Yu, Gu, Lezama, Huang, Schindler, Hornung, Birodkar, Yan, Chiu, et~al.]{videopoet}
Dan Kondratyuk, Lijun Yu, Xiuye Gu, Jos{\'e} Lezama, Jonathan Huang, Grant Schindler, Rachel Hornung, Vighnesh Birodkar, Jimmy Yan, Ming-Chang Chiu, et~al.
\newblock Videopoet: A large language model for zero-shot video generation.
\newblock \emph{arXiv preprint arXiv:2312.14125}, 2023.

\bibitem[Hong et~al.(2022)Hong, Ding, Zheng, Liu, and Tang]{cogvideo}
Wenyi Hong, Ming Ding, Wendi Zheng, Xinghan Liu, and Jie Tang.
\newblock Cogvideo: Large-scale pretraining for text-to-video generation via transformers.
\newblock \emph{arXiv preprint arXiv:2205.15868}, 2022.

\bibitem[Blattmann et~al.(2023{\natexlab{a}})Blattmann, Rombach, Ling, Dockhorn, Kim, Fidler, and Kreis]{blattmann2023align}
Andreas Blattmann, Robin Rombach, Huan Ling, Tim Dockhorn, Seung~Wook Kim, Sanja Fidler, and Karsten Kreis.
\newblock Align your latents: High-resolution video synthesis with latent diffusion models.
\newblock In \emph{Proceedings of the IEEE/CVF Conference on Computer Vision and Pattern Recognition}, pages 22563--22575, 2023{\natexlab{a}}.

\bibitem[Esser et~al.(2023)Esser, Chiu, Atighehchian, Granskog, and Germanidis]{esser2023structure}
Patrick Esser, Johnathan Chiu, Parmida Atighehchian, Jonathan Granskog, and Anastasis Germanidis.
\newblock Structure and content-guided video synthesis with diffusion models.
\newblock In \emph{Proceedings of the IEEE/CVF International Conference on Computer Vision}, pages 7346--7356, 2023.

\bibitem[Bar et~al.(2024)Bar, Zhou, Tran, Darrell, and LeCun]{nwm}
Amir Bar, Gaoyue Zhou, Danny Tran, Trevor Darrell, and Yann LeCun.
\newblock Navigation world models.
\newblock \emph{arXiv preprint arXiv:2412.03572}, 2024.

\bibitem[Hu et~al.(2023)Hu, Russell, Yeo, Murez, Fedoseev, Kendall, Shotton, and Corrado]{gaia}
Anthony Hu, Lloyd Russell, Hudson Yeo, Zak Murez, George Fedoseev, Alex Kendall, Jamie Shotton, and Gianluca Corrado.
\newblock Gaia-1: A generative world model for autonomous driving.
\newblock \emph{arXiv preprint arXiv:2309.17080}, 2023.

\bibitem[Valevski et~al.(2024)Valevski, Leviathan, Arar, and Fruchter]{gameengine}
Dani Valevski, Yaniv Leviathan, Moab Arar, and Shlomi Fruchter.
\newblock Diffusion models are real-time game engines.
\newblock \emph{arXiv preprint arXiv:2408.14837}, 2024.

\bibitem[Hafner et~al.(2020)Hafner, Lillicrap, Ba, and Norouzi]{dreamer}
Danijar Hafner, Timothy Lillicrap, Jimmy Ba, and Mohammad Norouzi.
\newblock Dream to control: Learning behaviors by latent imagination.
\newblock In \emph{International Conference on Learning Representations}, 2020.

\bibitem[Bruce et~al.(2024)Bruce, Dennis, Edwards, Parker-Holder, Shi, Hughes, Lai, Mavalankar, Steigerwald, Apps, et~al.]{genie}
Jake Bruce, Michael~D Dennis, Ashley Edwards, Jack Parker-Holder, Yuge Shi, Edward Hughes, Matthew Lai, Aditi Mavalankar, Richie Steigerwald, Chris Apps, et~al.
\newblock Genie: Generative interactive environments.
\newblock In \emph{Forty-first International Conference on Machine Learning}, 2024.

\bibitem[{OpenStreetMap contributors}(2017)]{openstreetmap2017}
{OpenStreetMap contributors}.
\newblock Planet dump retrieved from https://planet.osm.org.
\newblock \url{https://www.openstreetmap.org}, 2017.

\bibitem[Team(2024)]{chameleon}
Chameleon Team.
\newblock Chameleon: Mixed-modal early-fusion foundation models.
\newblock \emph{arXiv preprint arXiv:2405.09818}, 2024.

\bibitem[Kalchbrenner et~al.(2017{\natexlab{a}})Kalchbrenner, van~den Oord, Simonyan, Danihelka, Vinyals, Graves, and Kavukcuoglu]{vpnkalchbrenner17a}
Nal Kalchbrenner, A{\"a}ron van~den Oord, Karen Simonyan, Ivo Danihelka, Oriol Vinyals, Alex Graves, and Koray Kavukcuoglu.
\newblock Video pixel networks.
\newblock In Doina Precup and Yee~Whye Teh, editors, \emph{Proceedings of the 34th International Conference on Machine Learning}, volume~70 of \emph{Proceedings of Machine Learning Research}, pages 1771--1779. PMLR, 06--11 Aug 2017{\natexlab{a}}.
\newblock URL \url{https://proceedings.mlr.press/v70/kalchbrenner17a.html}.

\bibitem[Clark et~al.(2019)Clark, Donahue, and Simonyan]{dvdgan}
Aidan Clark, Jeff Donahue, and Karen Simonyan.
\newblock Efficient video generation on complex datasets.
\newblock \emph{CoRR}, abs/1907.06571, 2019.
\newblock URL \url{http://arxiv.org/abs/1907.06571}.

\bibitem[Finn et~al.(2016)Finn, Goodfellow, and Levine]{finn2016}
Chelsea Finn, Ian Goodfellow, and Sergey Levine.
\newblock Unsupervised learning for physical interaction through video prediction.
\newblock In \emph{Proceedings of the 30th International Conference on Neural Information Processing Systems}, NIPS'16, page 64–72, Red Hook, NY, USA, 2016. Curran Associates Inc.
\newblock ISBN 9781510838819.

\bibitem[Luc et~al.(2020)Luc, Clark, Dieleman, de~Las~Casas, Doron, Cassirer, and Simonyan]{trivdgan2020}
Pauline Luc, Aidan Clark, Sander Dieleman, Diego de~Las~Casas, Yotam Doron, Albin Cassirer, and Karen Simonyan.
\newblock Transformation-based adversarial video prediction on large-scale data.
\newblock \emph{CoRR}, abs/2003.04035, 2020.

\bibitem[Lotter et~al.(2017)Lotter, Kreiman, and Cox]{lotter2017deep}
William Lotter, Gabriel Kreiman, and David Cox.
\newblock Deep predictive coding networks for video prediction and unsupervised learning.
\newblock In \emph{International Conference on Learning Representations}, 2017.
\newblock URL \url{https://openreview.net/forum?id=B1ewdt9xe}.

\bibitem[Yan et~al.(2021)Yan, Zhang, Abbeel, and Srinivas]{yan2021videogpt}
Wilson Yan, Yunzhi Zhang, Pieter Abbeel, and Aravind Srinivas.
\newblock Videogpt: Video generation using vq-vae and transformers, 2021.

\bibitem[Blattmann et~al.(2023{\natexlab{b}})Blattmann, Rombach, Ling, Dockhorn, Kim, Fidler, and Kreis]{Blattmann2023AlignYL}
A.~Blattmann, Robin Rombach, Huan Ling, Tim Dockhorn, Seung~Wook Kim, Sanja Fidler, and Karsten Kreis.
\newblock Align your latents: High-resolution video synthesis with latent diffusion models.
\newblock \emph{2023 IEEE/CVF Conference on Computer Vision and Pattern Recognition (CVPR)}, pages 22563--22575, 2023{\natexlab{b}}.

\bibitem[Walker et~al.(2021)Walker, Razavi, and van~den Oord]{walker2021predicting}
Jacob~C Walker, Ali Razavi, and Aaron van~den Oord.
\newblock Predicting video with {VQVAE}, 2021.

\bibitem[Le~Moing et~al.(2021)Le~Moing, Ponce, and Schmid]{NEURIPS2021_757b505c}
Guillaume Le~Moing, Jean Ponce, and Cordelia Schmid.
\newblock Ccvs: Context-aware controllable video synthesis.
\newblock In M.~Ranzato, A.~Beygelzimer, Y.~Dauphin, P.S. Liang, and J.~Wortman Vaughan, editors, \emph{Advances in Neural Information Processing Systems}, volume~34, pages 14042--14055. Curran Associates, Inc., 2021.

\bibitem[H{\"o}ppe et~al.(2022)H{\"o}ppe, Mehrjou, Bauer, Nielsen, and Dittadi]{hoppe2022diffusion}
Tobias H{\"o}ppe, Arash Mehrjou, Stefan Bauer, Didrik Nielsen, and Andrea Dittadi.
\newblock Diffusion models for video prediction and infilling.
\newblock \emph{Transactions on Machine Learning Research}, 2022.
\newblock ISSN 2835-8856.

\bibitem[Singer et~al.(2023)Singer, Polyak, Hayes, Yin, An, Zhang, Hu, Yang, Ashual, Gafni, Parikh, Gupta, and Taigman]{singer2023makeavideo}
Uriel Singer, Adam Polyak, Thomas Hayes, Xi~Yin, Jie An, Songyang Zhang, Qiyuan Hu, Harry Yang, Oron Ashual, Oran Gafni, Devi Parikh, Sonal Gupta, and Yaniv Taigman.
\newblock Make-a-video: Text-to-video generation without text-video data.
\newblock In \emph{The Eleventh International Conference on Learning Representations}, 2023.

\bibitem[Ho et~al.(2022{\natexlab{a}})Ho, Chan, Saharia, Whang, Gao, Gritsenko, Kingma, Poole, Norouzi, Fleet, and Salimans]{ho2022imagen}
Jonathan Ho, William Chan, Chitwan Saharia, Jay Whang, Ruiqi Gao, Alexey Gritsenko, Diederik~P. Kingma, Ben Poole, Mohammad Norouzi, David~J. Fleet, and Tim Salimans.
\newblock Imagen video: High definition video generation with diffusion models, 2022{\natexlab{a}}.

\bibitem[Ho et~al.(2022{\natexlab{b}})Ho, Salimans, Gritsenko, Chan, Norouzi, and Fleet]{NEURIPS2022_39235c56}
Jonathan Ho, Tim Salimans, Alexey Gritsenko, William Chan, Mohammad Norouzi, and David~J Fleet.
\newblock Video diffusion models.
\newblock In S.~Koyejo, S.~Mohamed, A.~Agarwal, D.~Belgrave, K.~Cho, and A.~Oh, editors, \emph{Advances in Neural Information Processing Systems}, volume~35, pages 8633--8646. Curran Associates, Inc., 2022{\natexlab{b}}.

\bibitem[Yu et~al.(2023)Yu, Cheng, Sohn, Lezama, Zhang, Chang, Hauptmann, Yang, Hao, Essa, and Jiang]{10205485}
L.~Yu, Y.~Cheng, K.~Sohn, J.~Lezama, H.~Zhang, H.~Chang, A.~G. Hauptmann, M.~Yang, Y.~Hao, I.~Essa, and L.~Jiang.
\newblock Magvit: Masked generative video transformer.
\newblock In \emph{2023 IEEE/CVF Conference on Computer Vision and Pattern Recognition (CVPR)}, pages 10459--10469, Los Alamitos, CA, USA, jun 2023. IEEE Computer Society.
\newblock \doi{10.1109/CVPR52729.2023.01008}.

\bibitem[Bai et~al.(2024)Bai, Geng, Mangalam, Bar, Yuille, Darrell, Malik, and Efros]{lvm}
Yutong Bai, Xinyang Geng, Karttikeya Mangalam, Amir Bar, Alan~L Yuille, Trevor Darrell, Jitendra Malik, and Alexei~A Efros.
\newblock Sequential modeling enables scalable learning for large vision models.
\newblock In \emph{Proceedings of the IEEE/CVF Conference on Computer Vision and Pattern Recognition}, pages 22861--22872, 2024.

\bibitem[Kim et~al.(2021{\natexlab{a}})Kim, Philion, Torralba, and Fidler]{drivegan}
Seung~Wook Kim, Jonah Philion, Antonio Torralba, and Sanja Fidler.
\newblock Drivegan: Towards a controllable high-quality neural simulation.
\newblock In \emph{Proceedings of the IEEE/CVF Conference on Computer Vision and Pattern Recognition}, pages 5820--5829, 2021{\natexlab{a}}.

\bibitem[Sun et~al.(2020)Sun, Kretzschmar, Dotiwalla, Chouard, Patnaik, Tsui, Guo, Zhou, Chai, Caine, Vasudevan, Han, Ngiam, Zhao, Timofeev, Ettinger, Krivokon, Gao, Joshi, Zhang, Shlens, Chen, and Anguelov]{waymo}
Pei Sun, Henrik Kretzschmar, Xerxes Dotiwalla, Aurelien Chouard, Vijaysai Patnaik, Paul Tsui, James Guo, Yin Zhou, Yuning Chai, Benjamin Caine, Vijay Vasudevan, Wei Han, Jiquan Ngiam, Hang Zhao, Aleksei Timofeev, Scott Ettinger, Maxim Krivokon, Amy Gao, Aditya Joshi, Yu~Zhang, Jonathon Shlens, Zhifeng Chen, and Dragomir Anguelov.
\newblock Scalability in perception for autonomous driving: Waymo open dataset.
\newblock In \emph{2020 {IEEE/CVF} Conference on Computer Vision and Pattern Recognition, {CVPR} 2020, Seattle, WA, USA, June 13-19, 2020}, pages 2443--2451. Computer Vision Foundation / {IEEE}, 2020.
\newblock \doi{10.1109/CVPR42600.2020.00252}.
\newblock URL \url{https://openaccess.thecvf.com/content\_CVPR\_2020/html/Sun\_Scalability\_in\_Perception\_for\_Autonomous\_Driving\_Waymo\_Open\_Dataset\_CVPR\_2020\_paper.html}.

\bibitem[Caesar et~al.(2020)Caesar, Bankiti, Lang, Vora, Liong, Xu, Krishnan, Pan, Baldan, and Beijbom]{nuscenes}
Holger Caesar, Varun Bankiti, Alex~H. Lang, Sourabh Vora, Venice~Erin Liong, Qiang Xu, Anush Krishnan, Yu~Pan, Giancarlo Baldan, and Oscar Beijbom.
\newblock nuscenes: A multimodal dataset for autonomous driving.
\newblock In \emph{CVPR}, 2020.

\bibitem[Wilson et~al.(2021)Wilson, Qi, Agarwal, Lambert, Singh, Khandelwal, Pan, Kumar, Hartnett, Pontes, Ramanan, Carr, and Hays]{Argoverse2}
Benjamin Wilson, William Qi, Tanmay Agarwal, John Lambert, Jagjeet Singh, Siddhesh Khandelwal, Bowen Pan, Ratnesh Kumar, Andrew Hartnett, Jhony~Kaesemodel Pontes, Deva Ramanan, Peter Carr, and James Hays.
\newblock Argoverse 2: Next generation datasets for self-driving perception and forecasting.
\newblock In \emph{Proceedings of the Neural Information Processing Systems Track on Datasets and Benchmarks (NeurIPS Datasets and Benchmarks 2021)}, 2021.

\bibitem[Ha and Schmidhuber(2018)]{worldmodels}
David Ha and J\"{u}rgen Schmidhuber.
\newblock Recurrent world models facilitate policy evolution.
\newblock In \emph{Proceedings of the 32Nd International Conference on Neural Information Processing Systems}, NeurIPS'18, pages 2455--2467, 2018.

\bibitem[Oh et~al.(2015)Oh, Guo, Lee, Lewis, and Singh]{oh2015}
Junhyuk Oh, Xiaoxiao Guo, Honglak Lee, Richard Lewis, and Satinder Singh.
\newblock Action-conditional video prediction using deep networks in atari games.
\newblock In \emph{Proceedings of the 28th International Conference on Neural Information Processing Systems - Volume 2}, NIPS'15, page 2863–2871, Cambridge, MA, USA, 2015. MIT Press.

\bibitem[Nunes et~al.(2020)Nunes, Dehban, Moreno, and Santos-Victor]{actionsurvey}
Manuel~Serra Nunes, Atabak Dehban, Plinio Moreno, and José Santos-Victor.
\newblock Action-conditioned benchmarking of robotic video prediction models: a comparative study.
\newblock In \emph{2020 IEEE International Conference on Robotics and Automation (ICRA)}, pages 8316--8322, 2020.
\newblock \doi{10.1109/ICRA40945.2020.9196839}.

\bibitem[Hafner et~al.(2021)Hafner, Lillicrap, Norouzi, and Ba]{hafner2021mastering}
Danijar Hafner, Timothy~P Lillicrap, Mohammad Norouzi, and Jimmy Ba.
\newblock Mastering atari with discrete world models.
\newblock In \emph{International Conference on Learning Representations}, 2021.

\bibitem[Micheli et~al.(2023)Micheli, Alonso, and Fleuret]{micheli2023transformers}
Vincent Micheli, Eloi Alonso, and Fran{\c{c}}ois Fleuret.
\newblock Transformers are sample-efficient world models.
\newblock In \emph{The Eleventh International Conference on Learning Representations}, 2023.

\bibitem[Robine et~al.(2023)Robine, H{\"o}ftmann, Uelwer, and Harmeling]{robine2023transformerbased}
Jan Robine, Marc H{\"o}ftmann, Tobias Uelwer, and Stefan Harmeling.
\newblock Transformer-based world models are happy with 100k interactions.
\newblock In \emph{The Eleventh International Conference on Learning Representations}, 2023.

\bibitem[Kim et~al.(2020)Kim, Zhou, Philion, Torralba, and Fidler]{Kim_2020_CVPR}
Seung~Wook Kim, Yuhao Zhou, Jonah Philion, Antonio Torralba, and Sanja Fidler.
\newblock Learning to simulate dynamic environments with gamegan.
\newblock In \emph{Proceedings of the IEEE/CVF Conference on Computer Vision and Pattern Recognition (CVPR)}, June 2020.

\bibitem[Kim et~al.(2021{\natexlab{b}})Kim, Philion, Torralba, and Fidler]{Kim_2021_CVPR}
Seung~Wook Kim, Jonah Philion, Antonio Torralba, and Sanja Fidler.
\newblock Drivegan: Towards a controllable high-quality neural simulation.
\newblock In \emph{Proceedings of the IEEE/CVF Conference on Computer Vision and Pattern Recognition (CVPR)}, pages 5820--5829, June 2021{\natexlab{b}}.

\bibitem[Bamford and Lucas(2020)]{bamfordnge2020}
Chris Bamford and Simon~M. Lucas.
\newblock Neural game engine: Accurate learning ofgeneralizable forward models from pixels.
\newblock In \emph{Conference on Games}, 2020.

\bibitem[Chiappa et~al.(2017)Chiappa, Racaniere, Wierstra, and Mohamed]{chiappa2017recurrent}
Silvia Chiappa, S{\'e}bastien Racaniere, Daan Wierstra, and Shakir Mohamed.
\newblock Recurrent environment simulators.
\newblock In \emph{International Conference on Learning Representations}, 2017.

\bibitem[Pan et~al.(2022)Pan, Zhu, Wang, and Yang]{NEURIPS2022_9316769a}
Minting Pan, Xiangming Zhu, Yunbo Wang, and Xiaokang Yang.
\newblock Iso-dream: Isolating and leveraging noncontrollable visual dynamics in world models.
\newblock In S.~Koyejo, S.~Mohamed, A.~Agarwal, D.~Belgrave, K.~Cho, and A.~Oh, editors, \emph{Advances in Neural Information Processing Systems}, volume~35, pages 23178--23191. Curran Associates, Inc., 2022.

\bibitem[Eslami et~al.(2018)Eslami, Rezende, Besse, Viola, Morcos, Garnelo, Ruderman, Rusu, Danihelka, Gregor, Reichert, Buesing, Weber, Vinyals, Rosenbaum, Rabinowitz, King, Hillier, Botvinick, Wierstra, Kavukcuoglu, and Hassabis]{scienceaar6170}
S.~M.~Ali Eslami, Danilo~Jimenez Rezende, Frederic Besse, Fabio Viola, Ari~S. Morcos, Marta Garnelo, Avraham Ruderman, Andrei~A. Rusu, Ivo Danihelka, Karol Gregor, David~P. Reichert, Lars Buesing, Theophane Weber, Oriol Vinyals, Dan Rosenbaum, Neil Rabinowitz, Helen King, Chloe Hillier, Matt Botvinick, Daan Wierstra, Koray Kavukcuoglu, and Demis Hassabis.
\newblock Neural scene representation and rendering.
\newblock \emph{Science}, 360\penalty0 (6394):\penalty0 1204--1210, 2018.
\newblock \doi{10.1126/science.aar6170}.

\bibitem[Wang et~al.(2023)Wang, Zhu, Huang, Chen, Zhu, and Lu]{drivedreamer}
Xiaofeng Wang, Zheng Zhu, Guan Huang, Xinze Chen, Jiagang Zhu, and Jiwen Lu.
\newblock Drivedreamer: Towards real-world-driven world models for autonomous driving.
\newblock \emph{arXiv preprint arXiv:2309.09777}, 2023.

\bibitem[Gao et~al.(2024)Gao, Yang, Chen, Chitta, Qiu, Geiger, Zhang, and Li]{vista}
Shenyuan Gao, Jiazhi Yang, Li~Chen, Kashyap Chitta, Yihang Qiu, Andreas Geiger, Jun Zhang, and Hongyang Li.
\newblock Vista: A generalizable driving world model with high fidelity and versatile controllability.
\newblock \emph{arXiv preprint arXiv:2405.17398}, 2024.

\bibitem[Van Den~Oord et~al.(2016)Van Den~Oord, Kalchbrenner, and Kavukcuoglu]{oord2016pixelrnn}
A{\"a}ron Van Den~Oord, Nal Kalchbrenner, and Koray Kavukcuoglu.
\newblock Pixel recurrent neural networks.
\newblock In \emph{International conference on machine learning}, pages 1747--1756. PMLR, 2016.

\bibitem[Kalchbrenner et~al.(2017{\natexlab{b}})Kalchbrenner, Oord, Simonyan, Danihelka, Vinyals, Graves, and Kavukcuoglu]{kalchbrenner2017video}
Nal Kalchbrenner, A{\"a}ron Oord, Karen Simonyan, Ivo Danihelka, Oriol Vinyals, Alex Graves, and Koray Kavukcuoglu.
\newblock Video pixel networks.
\newblock In \emph{International Conference on Machine Learning}, pages 1771--1779. PMLR, 2017{\natexlab{b}}.

\bibitem[Gupta et~al.(2023)Gupta, Tian, Zhang, Wu, Mart{\'\i}n-Mart{\'\i}n, and Fei-Fei]{gupta2023maskvit}
Agrim Gupta, Stephen Tian, Yunzhi Zhang, Jiajun Wu, Roberto Mart{\'\i}n-Mart{\'\i}n, and Li~Fei-Fei.
\newblock Maskvit: Masked visual pre-training for video prediction.
\newblock In \emph{The Eleventh International Conference on Learning Representations}, 2023.
\newblock URL \url{https://openreview.net/forum?id=QAV2CcLEDh}.

\bibitem[Deng et~al.(2025)Deng, Pan, Diao, Luo, Cui, Lu, Shan, Qi, and Wang]{deng2025autoregressive}
Haoge Deng, Ting Pan, Haiwen Diao, Zhengxiong Luo, Yufeng Cui, Huchuan Lu, Shiguang Shan, Yonggang Qi, and Xinlong Wang.
\newblock Autoregressive video generation without vector quantization.
\newblock In \emph{The Thirteenth International Conference on Learning Representations}, 2025.
\newblock URL \url{https://openreview.net/forum?id=JE9tCwe3lp}.

\bibitem[Gu et~al.(2025)Gu, Mao, and Shou]{gu2025long}
Yuchao Gu, Weijia Mao, and Mike~Zheng Shou.
\newblock Long-context autoregressive video modeling with next-frame prediction.
\newblock \emph{arXiv preprint arXiv:2503.19325}, 2025.

\bibitem[Yu et~al.(2020)Yu, Chen, Wang, Xian, Chen, Liu, Madhavan, and Darrell]{bdd100k}
Fisher Yu, Haofeng Chen, Xin Wang, Wenqi Xian, Yingying Chen, Fangchen Liu, Vashisht Madhavan, and Trevor Darrell.
\newblock Bdd100k: A diverse driving dataset for heterogeneous multitask learning.
\newblock In \emph{Proceedings of the IEEE/CVF conference on computer vision and pattern recognition}, pages 2636--2645, 2020.

\bibitem[Kesten et~al.(2019)Kesten, Usman, Houston, Pandya, Nadhamuni, Ferreira, Yuan, Low, Jain, Ondruska, Omari, Shah, Kulkarni, Kazakova, Tao, Platinsky, Jiang, and Shet]{Woven}
R.~Kesten, M.~Usman, J.~Houston, T.~Pandya, K.~Nadhamuni, A.~Ferreira, M.~Yuan, B.~Low, A.~Jain, P.~Ondruska, S.~Omari, S.~Shah, A.~Kulkarni, A.~Kazakova, C.~Tao, L.~Platinsky, W.~Jiang, and V.~Shet.
\newblock Woven planet perception dataset 2020.
\newblock \url{https://woven.toyota/en/perception-dataset}, 2019.

\bibitem[Touvron et~al.(2023)Touvron, Lavril, Izacard, Martinet, Lachaux, Lacroix, Rozi{\`e}re, Goyal, Hambro, Azhar, et~al.]{llama}
Hugo Touvron, Thibaut Lavril, Gautier Izacard, Xavier Martinet, Marie-Anne Lachaux, Timoth{\'e}e Lacroix, Baptiste Rozi{\`e}re, Naman Goyal, Eric Hambro, Faisal Azhar, et~al.
\newblock Llama: Open and efficient foundation language models.
\newblock \emph{arXiv preprint arXiv:2302.13971}, 2023.

\end{thebibliography}
}

\newpage \clearpage
\appendix

\section{Limitations} 
\label{sec:limitations-future}




While TARDIS and STRIDE advance spatiotemporal world modeling, they inherit constraints common to complex AI systems. We discuss some of these limitations to guide responsible development and highlight promising research directions.

\subsection{STRIDE Dataset Limitations}

The current STRIDE dataset focuses on urban environments in a single geographic region (San Mateo County, CA), limiting applicability to rural areas and diverse city layouts. While spanning 16 years, temporal sampling remains sparse post-2020, potentially underrepresenting rapid infrastructure changes. Data quality considerations include projection artifacts from panoramic stitching and residual privacy-sensitive content despite blurring protocols. Future iterations could expand coverage through multi-city partnerships and integrate complementary modalities like LIDAR for enhanced 3D consistency. In addition, the base street view imagery temporal resolution is at minimum one month, which is a limiting factor if intending to learn second, minute, hour or day interactions. Live driving videos are a better source of high temporal frequency data.


\subsection{TARDIS Model Limitations}

\begin{figure*}[h]
    \centering
    \includegraphics[width=1\linewidth]{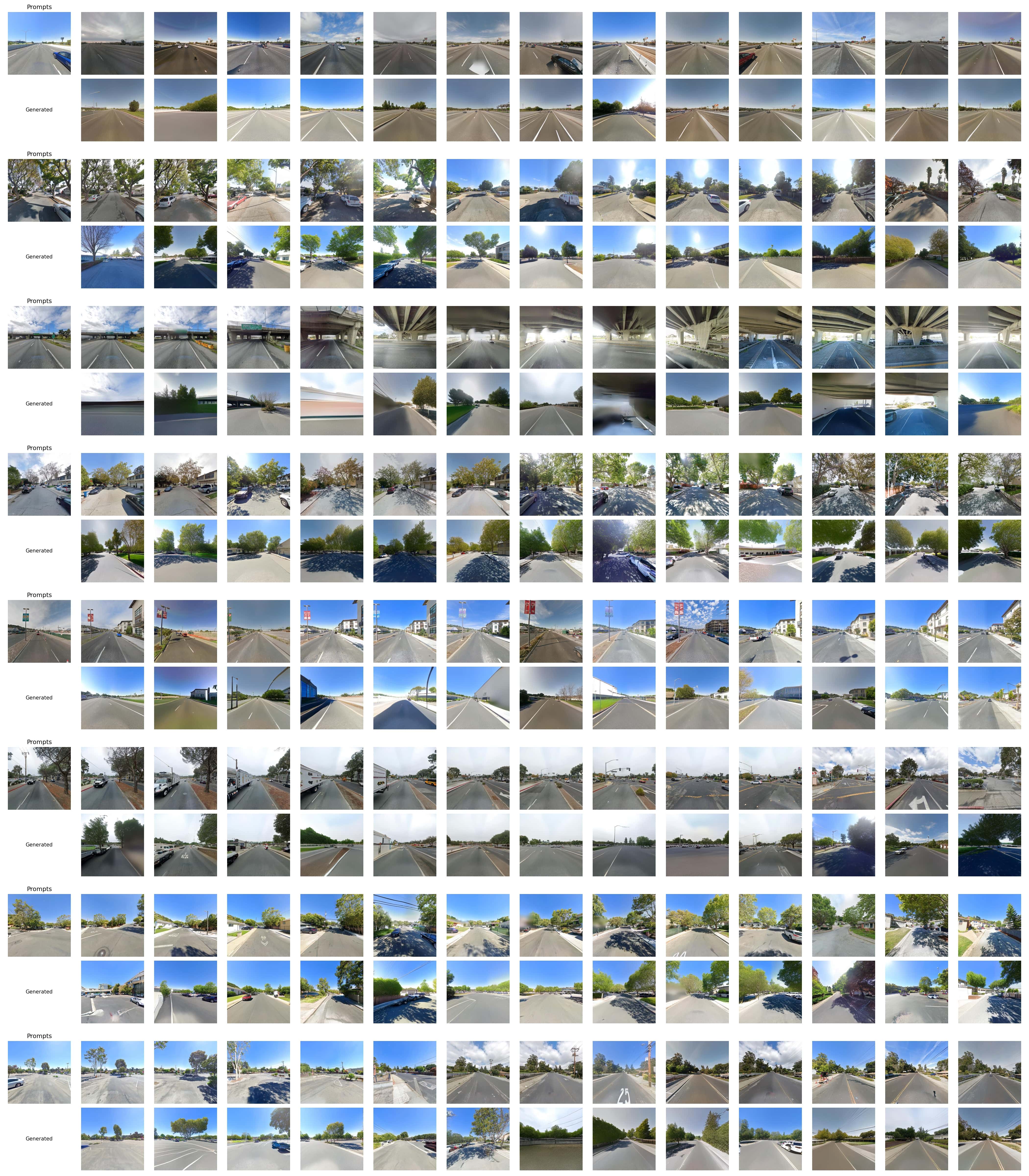}
    \caption{\textbf{Qualitative TARDIS Samples.} As mentioned on the data section, STRIDE lacks quality data around overpasses, which leads to degraded performance on row 3.}
    \label{fig:tardis-stack}
\end{figure*}

TARDIS's 1B-parameter architecture achieves strong spatiotemporal coherence but requires substantial compute resources, hindering real-time deployment on edge devices. While the model generalizes well within STRIDE's geographic scope, performance degrades in out-of-data distribution regions lacking similar road network topologies (c.f.~\cref{fig:tardis-stack}). The fixed 16K context window constrains extremely long-term temporal reasoning beyond 50-year projections. Promising future directions include: Modular architectures separating spatial/temporal pathways, cross-modal distillation to smaller foundation models, and hybrid neural-symbolic reasoning for out-of-data distribution generalization.

\section{Broader Impacts}
\label{sec:broader-impacts}

This work advances AI's capacity to model dynamic real-world environments, enabling applications in urban planning, disaster response, and climate resilience. By unifying spatial and temporal reasoning, it supports sustainable infrastructure development and proactive environmental monitoring. Future extensions could democratize access to high-fidelity simulations for global communities, fostering equitable resource management and adaptive policymaking. Ethical deployment frameworks ensure privacy preservation while leveraging geospatial insights. As spatiotemporal foundation models evolve, they hold promise for next-next generation environments built through responsible innovation.

\section{Related Work Discussion}
\label{sec:related}

\paragraph{Generative Video Models}
Recently, there has been tremendous progress with Generative Video Models~\cite{sora, videopoet, cogvideo, blattmann2023align, esser2023structure,vpnkalchbrenner17a, dvdgan, finn2016, trivdgan2020, lotter2017deep, yan2021videogpt, Blattmann2023AlignYL, walker2021predicting, NEURIPS2021_757b505c, hoppe2022diffusion, singer2023makeavideo, ho2022imagen, NEURIPS2022_39235c56, 10205485,lvm}, which typically condition on initial frames (or text) and predict the remaining frames in a video.
Frame prediction can be regarded as a special form of
generation, leveraging past observations to anticipate future frames~\cite{drivegan,chameleon}.

\paragraph{Open Datasets} Prior works have introduce world datasets for perception and localization to help solve some problems in the autonomous driving space.  Previous datasets like the Waymo Open Dataset~\cite{waymo}, nuScenes~\cite{nuscenes}, and Argoverse 2~\cite{Argoverse2} offer large-scale sensor data including LiDAR, camera feeds, and/or maps for autonomous driving research. However, these datasets primarily focus on spatial scene understanding for driving tasks and lack longitudinal temporal data capturing environmental changes over time needed for developing models capable of reasoning over long-term environmental dynamics.

\paragraph{Generative World Models}
The goal of World Models~\cite{worldmodels, oh2015}, is to simulate
the environment. Generative World Models can be considered a class of World Models which enable next-frame prediction that is conditioned on action inputs \cite{actionsurvey, dreamer, hafner2021mastering, micheli2023transformers, robine2023transformerbased, Kim_2020_CVPR, Kim_2021_CVPR, bamfordnge2020, chiappa2017recurrent, NEURIPS2022_9316769a, scienceaar6170}. World Models are crucial for agents to understand and interact with their environments, enabling prediction of future outcomes and informed decision-making.

In autonomous driving context, DriveGAN~\cite{drivegan} learns to simulate a driving scenario with vehicle control signals as its input. GAIA-1~\cite{gaia} and DriveDreamer~\cite{drivedreamer} further extend to action conditional diffusion models, enhancing the controllability and realism of generated videos. Vista~\cite{vista} further demonstrates the high-fidelity generation effect.

\paragraph{Auto-Regressive Modeling}
Auto-regressive models have driven significant advances in sequential data generation. PixelRNN~\cite{oord2016pixelrnn} pioneered autoregressive image generation using LSTMs to model pixel dependencies. This was extended to video with Video Pixel Networks~\cite{kalchbrenner2017video}, which captured spatiotemporal dependencies through 4D convolution chains. 
MaskViT~\cite{gupta2023maskvit} extended masked visual modeling to video prediction via spatial and spatiotemporal window attention. The emergence of non-quantized autoregressive models like NOVA~\cite{deng2025autoregressive} eliminated vector quantization bottlenecks through continuous-valued frame prediction. Recent work on long-context modeling~\cite{gu2025long} introduced Frame AutoRegressive (FAR) architectures with multi-level KV caching achieving frontier vision capabilities.
While these works demonstrated impressive sequence modeling capabilities, they primarily treat time as a passive prediction dimension rather than an actively navigable space.

While existing approaches model spatial dynamics well, they often treat time only as a prediction dimension. In contrast, in TARDIS, we make time actively navigable, combining spatial and temporal navigation into a unified process that enables real-time interactive verification.
Furthermore, in contrast to prior works limited to static snapshots or short-term observations, STRIDE enables holistic environment interaction through spatiotemporal permutations and active navigation. 
STRIDE introduces a composable, spatiotemporally rich dataset that explicitly models environmental changes across months and years, enabling agents not only to perceive but also to plan and adapt over extended temporal horizons. This facilitates research into long-horizon planning, time-aware localization, and embodied reasoning in dynamically evolving real-world environments.

STRIDE and TARDIS thus bridge the critical gap between passive scene understanding (BDD100K~\cite{bdd100k}) and dynamic world modeling (Woven perception dataset~\cite{Woven}), establishing a new paradigm for agent-centric spatiotemporal learning.


\begin{figure}[H]
    \centering
    \includegraphics[width=0.67\linewidth]{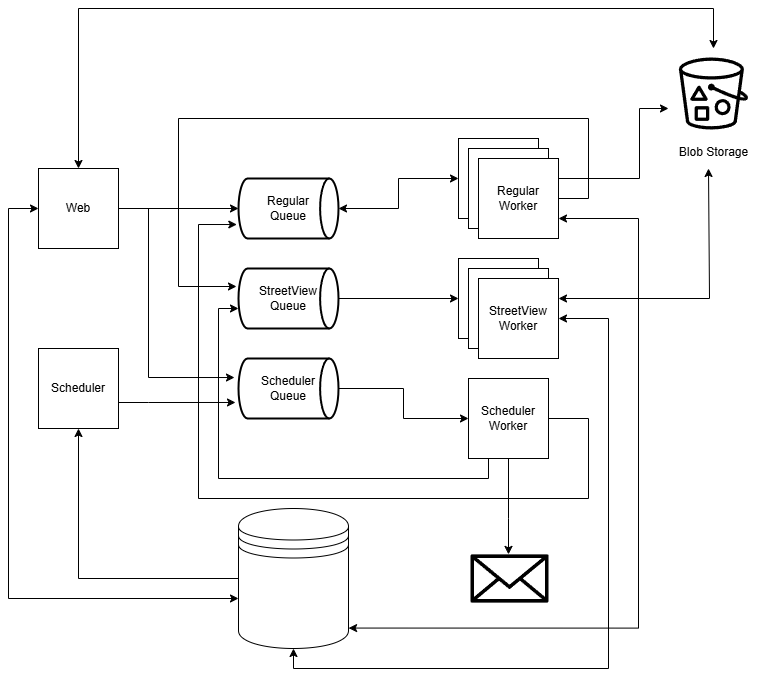}
    \caption{\textbf{General overview of the architecture for the distributed data aggregator.} We deployed a simple Kubernetes application where we independently scaled each worker to accommodate the desired throughput for Google StreetView images through the use of message brokers and queues. Decoupling image metadata from the binary allowed us to analyze the dataset before packaging and reexporting in the desired format (i.e. as a graph of interconnected nodes).}
    \label{fig:scraper-architecture}
\end{figure}





\section{Architecture and Implementation Details}
\label{sec:tardis-architecture}


Motivated by the sequential modeling, and the inherent sequential structure of spatio-temporal navigation model, we implement the following architecture for TARDIS:

\begin{table}[h!]
    \centering
        \begin{tabular}{lrrrr}\toprule
            &\textbf{Precision} & \textbf{Total Tokens} & \textbf{Possible Expressed Range}\\\midrule
            \textbf{Image} & N/A & 8192 & N/A \\
            \textbf{Latitude} & $1e^{-5}$ & 6723 & 37.50555, 37.57277 \\
            \textbf{Longitude} & $1e^{-5}$ & 10000 & -122.34916, -122.249168 \\
            \textbf{Month} & 1 & 12 & 1, 12 \\
            \textbf{Year} & 1 & 31 & 2000, 2030 \\
            \textbf{Distance} & 0.1 & 501 & 0.0, 50.0 \\
            \textbf{Heading} & 0.1 & 3601 & 0.0, 359.9 \\
            \textbf{$\Delta$ Month} & 1 & 12 & 0, 11 \\
            \textbf{$\Delta$ Year} & 1 & 61 & -30, 30
            \\\bottomrule
        \end{tabular}
    \vspace{0.5em}
    \caption{\textbf{Tokenization.} We dynamically tokenize different modalities with ranges and precision according to the size and resolution of the target dataset. Including special tokens, total vocabulary size is 29163.}
    \label{tokenization}
\end{table}

\paragraph{Image Tokenization} For image tokenization, we leverage a recently introduced VQGAN image tokenizer \cite{chameleon}. The image vocabulary size is $8192$ with each image being represented as $1024$ tokens. When decoding back into RGB space, the generated images are $512$x$512$ pixels in size.

\paragraph{Other-modality Tokenization} For all other modalities within $S$ and $A$, this is to say: latitude, longitude, month, year, distance, heading, month change and year chage, we tokenize each numerical value into discrete token bins. In order to cover the necessary range and resolution, we dynamically allocate more or fewer tokens based on the STRIDE dataset. For statistics on token assignment, value range and resolution see \textbf{Table \ref{tokenization}}.

We motivate the decision to tokenize into discrete categorical bins since the output from the model naturally forms a probabilistic distribution which we can directly sample from, enabling flexible conditional generation.

\paragraph{Base Architecture} We adopt a large transformer architecture with 1 billion parameters following the LLaMA~\cite{llama} design, with a effective batch size of 1024. Context length is 16K. For the rest of the hyperparameter design we follow~\cite{lvm}. We choose this as our base architecture as it is one of the most commonly used open-source LLM architectures. Since our task is significantly different from the original work, we train from scratch.


\section{Computational Resources}

\paragraph{Dataset Generation} Constructing the STRIDE dataset was done using the following hardware:
\begin{itemize}
    \item GPUs: N/A
    \item CPU: 8v Core Intel Ice Lake
    \item RAM: 64GB
    \item Storage: 512GB SSD
\end{itemize}

The dataset cluster consisted of a managed Kubernetes cluster with fewer than 10 nodes that hosted around 80 workers and ran for just under 2 days. The projection and tokenization of our arranged image dataset took around 734 days of wall time, or about 16 hours on our 128 VM cluster with the above hardware.

\paragraph{Training} All training experiments were conducted using the following hardware:

\begin{itemize}
    \item GPU: 32 NVIDIA H100 80GB
    \item CPU: 208v Core Intel Xeon
    \item RAM: 1.8TB
    \item Storage: 88GB SSD
\end{itemize}

The total compute cost for training TARDIS 1B was approximately 23,360 GPU hours or about a month on our 32 accelerator cluster. We believe this amount of training time was not enough to fully converge TARDIS.

\paragraph{Testing} Evaluation experiments were conducted using the following hardware:

\begin{itemize}
    \item GPU: NVIDIA A100 40GB
    \item CPU: 12v Core Intel Xeon
    \item RAM: 84GB
    \item Storage: 112GB SSD
\end{itemize}

The total compute cost for evaluating TARDIS on the test set data was approximately 250 GPU hours.




\section{Additional Image Generation Results}
\label{sec:appen-additional-results}

\begin{figure}[H]
    \centering
    \includegraphics[width=\linewidth, trim=0 1115 0 0, clip]{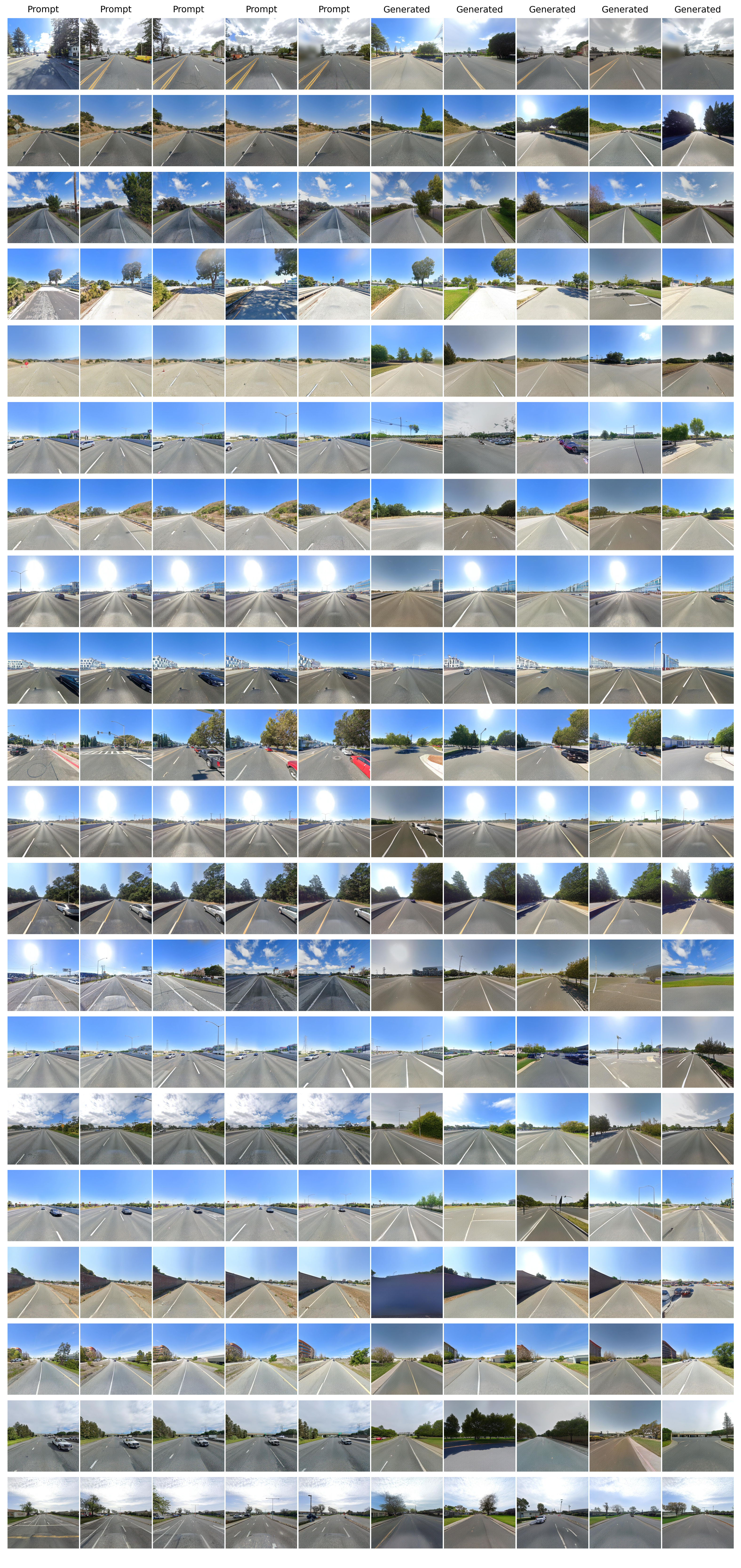}
    \caption{\textbf{Autoregressive TARDIS Generation.} In this example, we prompt TARDIS with five real images (left) and allow the model to extrapolate the next five frames without any additional supervision (right).}
    \label{fig:cropped-tardis-samples}
\end{figure}

\begin{figure}[b!]
    \centering
    \includegraphics[width=\linewidth]{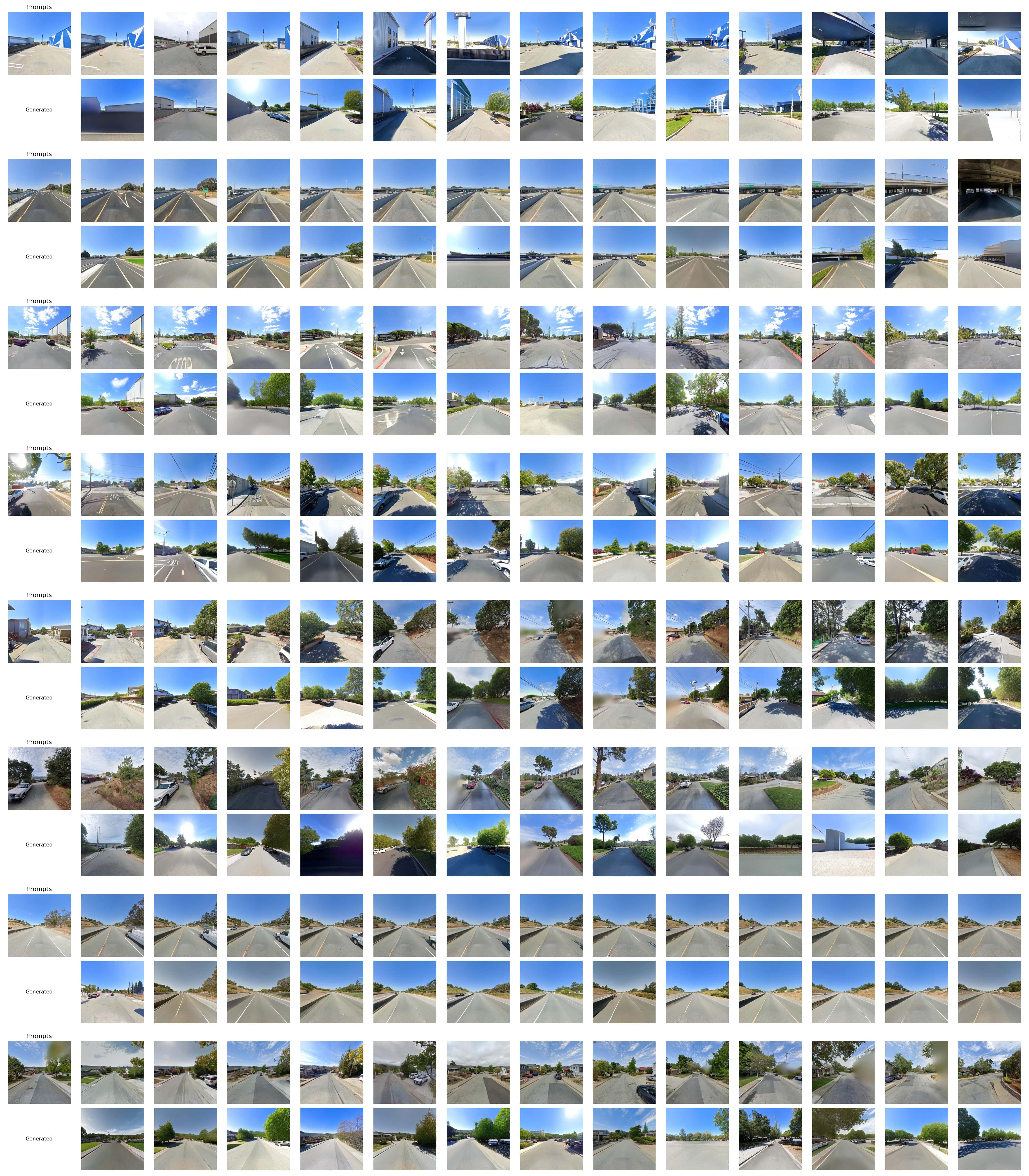}
    \caption{\textbf{Additional Qualitative TARDIS Samples \#1}}
    \label{fig:additional-qualitative-tardis-samples-4}
\end{figure}

\begin{figure}
    \centering
    \includegraphics[width=\linewidth]{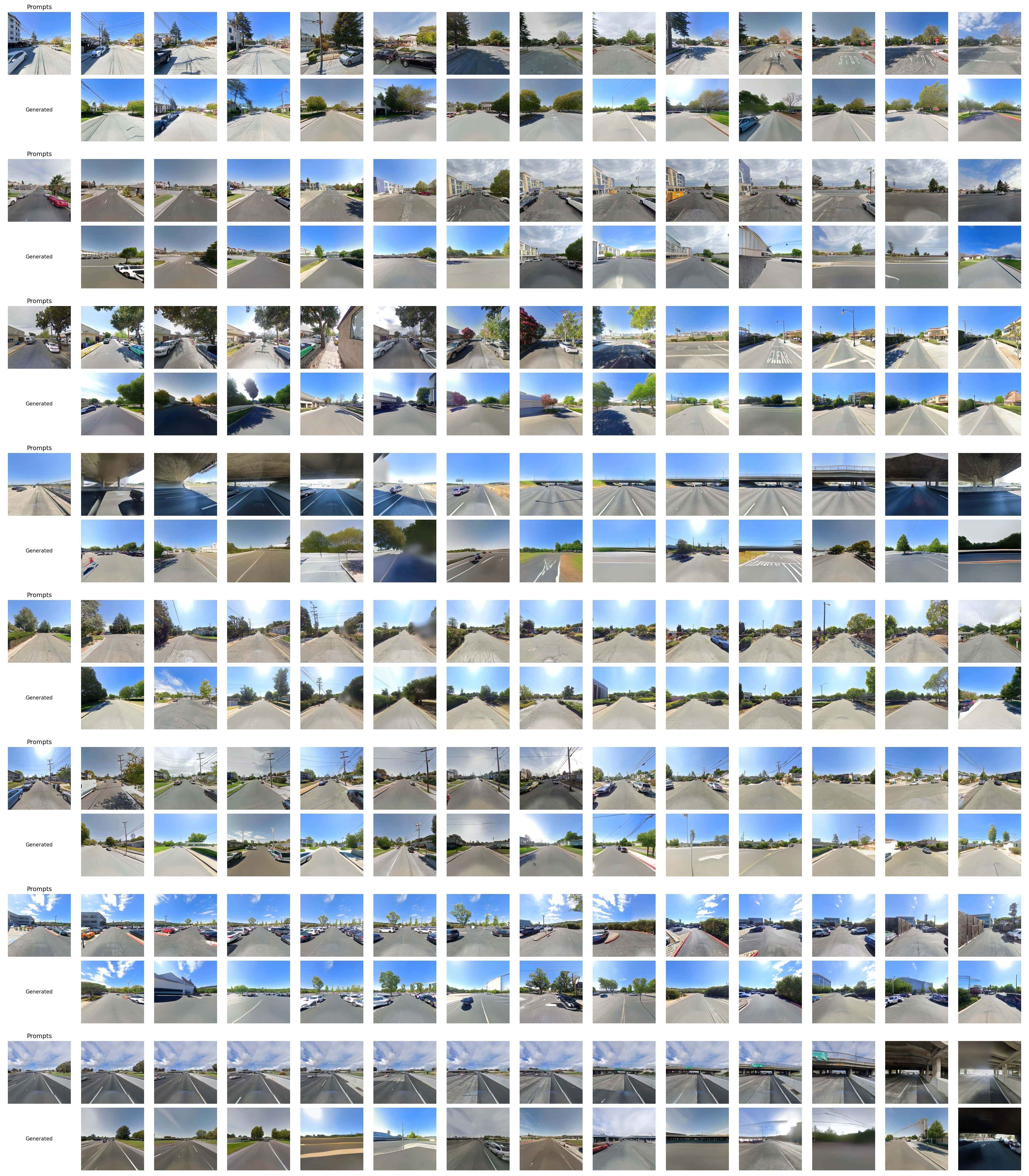}
    \caption{\textbf{Additional Qualitative TARDIS Samples \#2}}
    \label{fig:additional-qualitative-tardis-samples-2}
\end{figure}

\begin{figure}
    \centering
    \includegraphics[width=\linewidth]{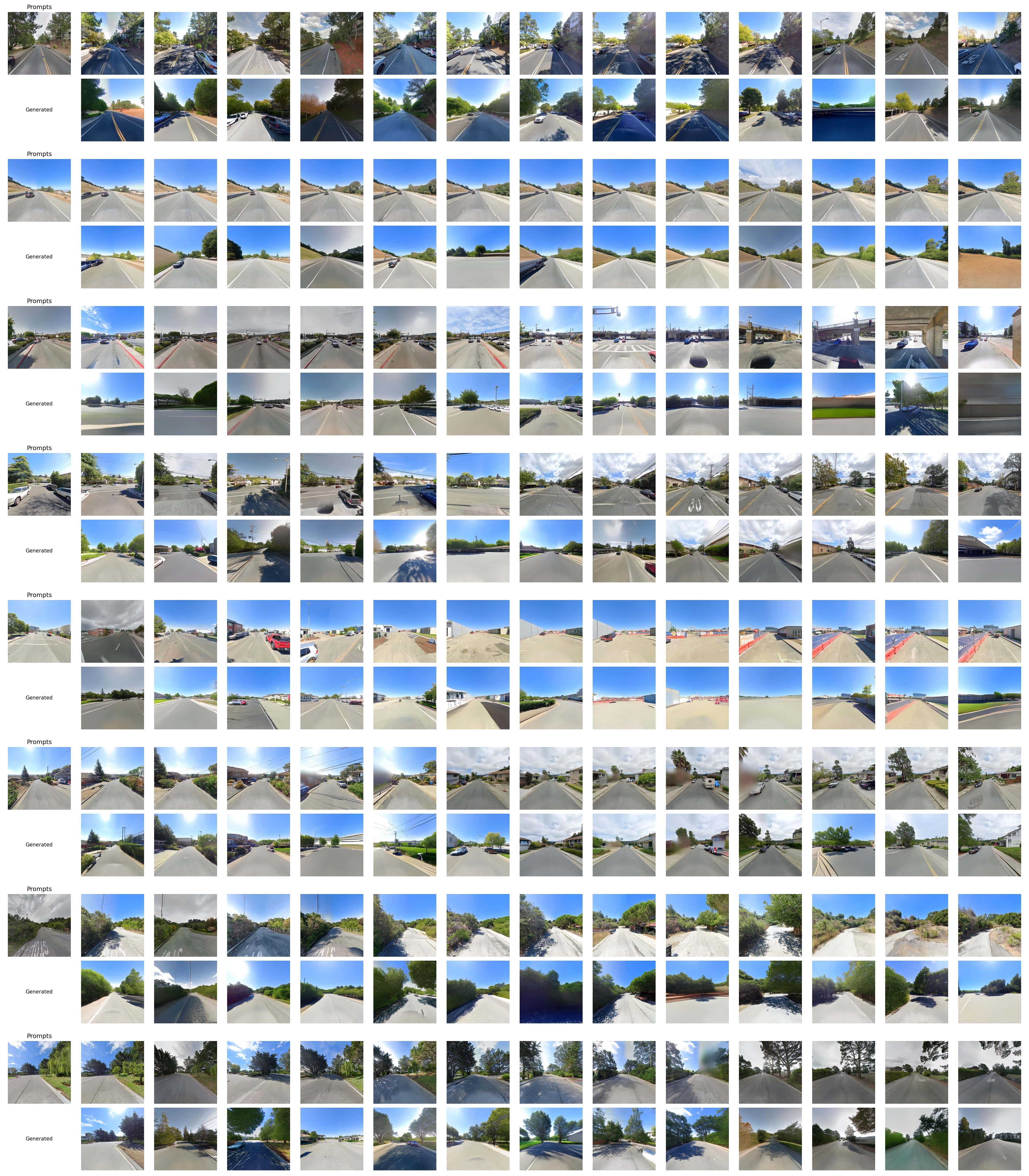}
    \caption{\textbf{Additional Qualitative TARDIS Samples \#3}}
    \label{fig:additional-qualitative-tardis-samples-3}
\end{figure}

\begin{figure}
    \centering
    \includegraphics[width=\linewidth]{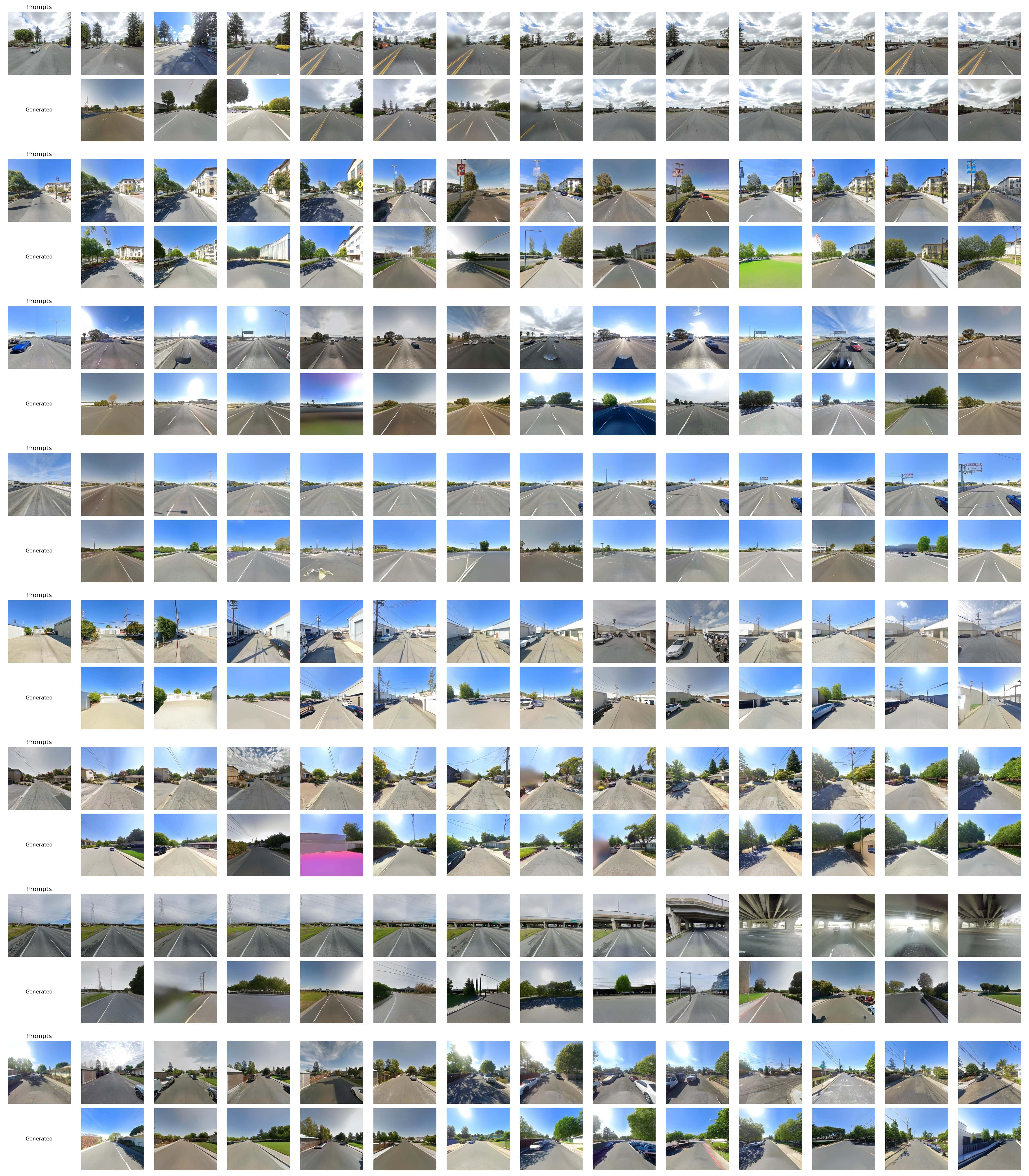}
    \caption{\textbf{Additional Qualitative TARDIS Samples \#4}}
    \label{fig:additional-qualitative-tardis-samples-1}
\end{figure}

\begin{figure}
    \centering
    \includegraphics[width=\linewidth]{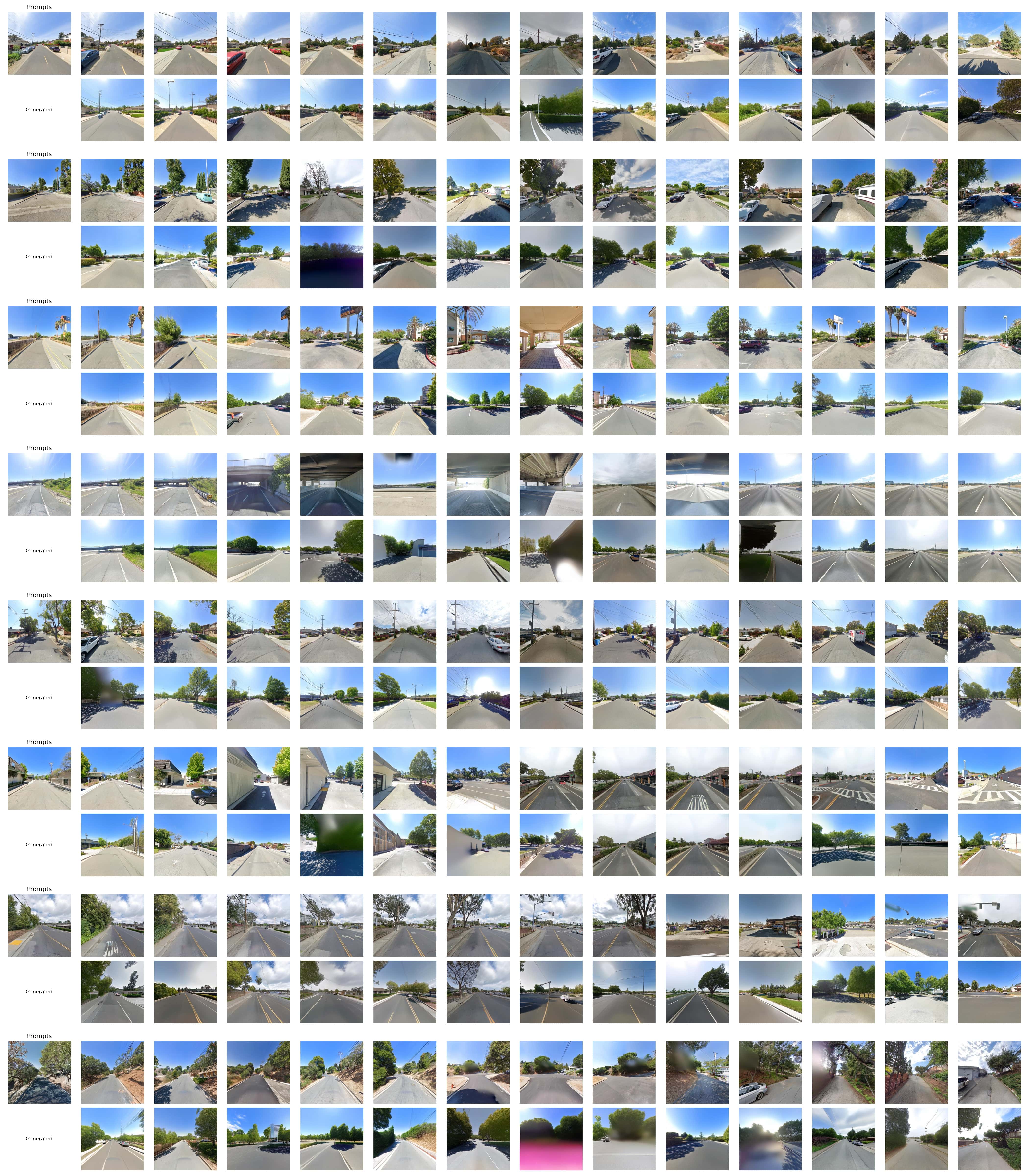}
    \caption{\textbf{Additional Qualitative TARDIS Samples \#5}}
    \label{fig:additional-qualitative-tardis-samples-5}
\end{figure}

We generate additional images test set images with TARDIS and show the results in large plots. On Figure \ref{fig:cropped-tardis-samples} we prompt TARDIS with five real image samples, then allow the model to extrapolate freely the next five observations. Figures \ref{fig:additional-qualitative-tardis-samples-4},\ref{fig:additional-qualitative-tardis-samples-2},\ref{fig:additional-qualitative-tardis-samples-3},\ref{fig:additional-qualitative-tardis-samples-1} and \ref{fig:additional-qualitative-tardis-samples-5} contain results where we generate the $n$-th image by prompting with real data up to step $n-1$, similar to what the model might experience in a live inference scenario.

\clearpage
\section{Additional Georeferencing Results}

\begin{figure}[H]
\centering
    \includegraphics[width=\linewidth]
    {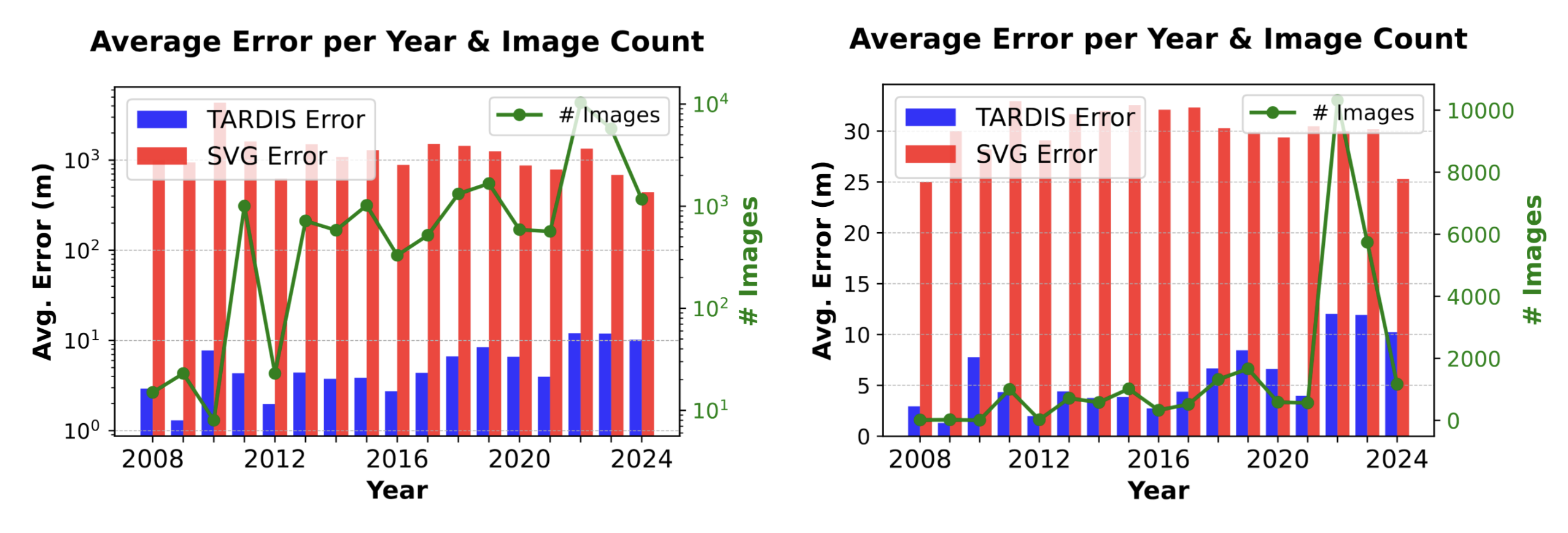}
    \caption{\textbf{Left:} Average Error Per Year of TARDIS vs SVG (full search space). \textbf{Right:} Average Error Per Year of TARDIS vs SVG with 50 Meter Search Space. These plots describe the fraction of predictions within some distance threshold.}
    \label{fig:tardis_svg_sidexside}
\end{figure}

\begin{figure}[H]
    \centering
    \includegraphics[width=0.6\linewidth]{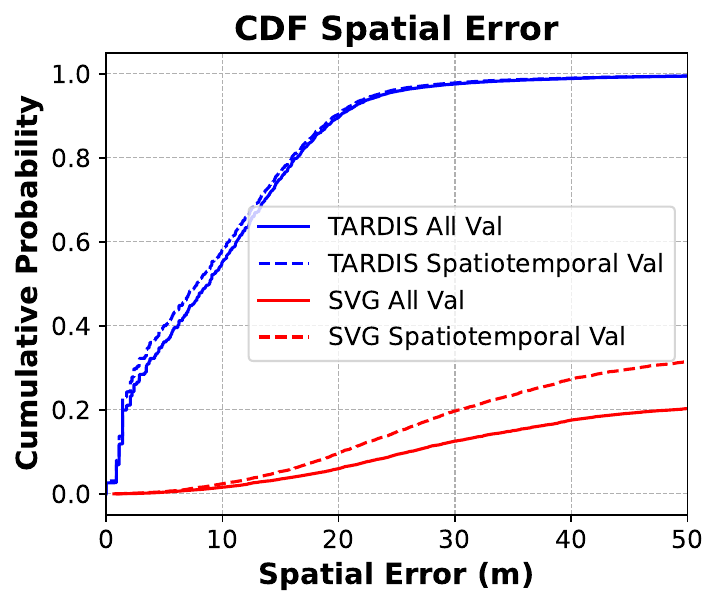}
    \caption{\textbf{CDF TARDIS vs SVG:} This describes the fraction of predictions within some distance threshold, we find TARDIS error to be within 20m 90\% of the time, while SVG is within this error at about a 10\% rate.}
    \label{fig:svg_tardis_cdf}
\end{figure}

We measure the error distribution comparison between TARDIS and SVG with two different search space settings for SVG (full area and 50m) on Figure \ref{fig:tardis_svg_sidexside}. Additionally, we plot the cumulative distribution function for TARDIS and SVG predictions, finding our method's coordinate prediction error to be within 20 meters for 90\% of predictions, while SVG reaching this error distance at a rate of about 10\%. This is visualized on Figure \ref{fig:svg_tardis_cdf}.
\begin{figure}[H]
\centering
    \includegraphics[width=\linewidth]{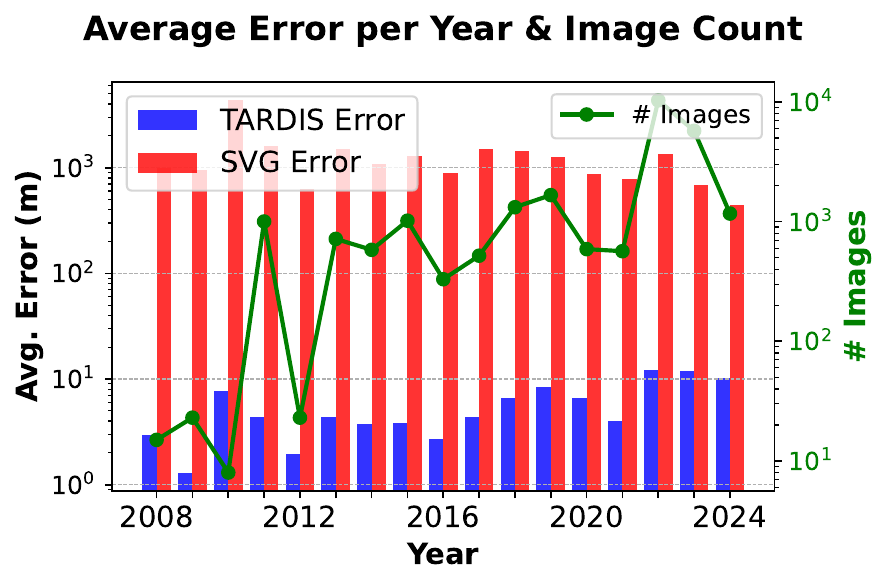}
    \caption{}
    \label{fig:svg_tardis_per_year}
\end{figure}

\section{Additional Self-Control Results}
\begin{figure}[H]
\centering
    \includegraphics[width=0.7\linewidth]{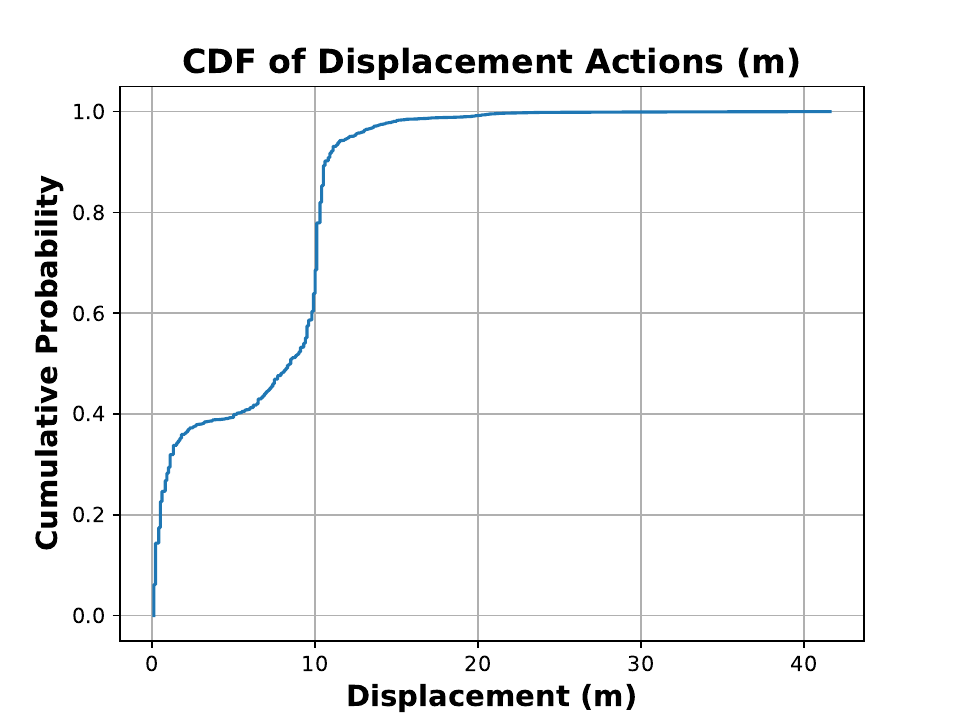}
    \caption{\textbf{TARDIS Self-Control Action Magnitude CDF.} Self-generated actions actions move at a magnitude of 5m or above 60\% of the time.}
    \label{fig:cdf-of-spatial-action}
\end{figure}

We observe the displacement actions of the model predictions to follow an interesting distribution. About 20\% of actions are of a magnitude of 1 meter or less, we believe this is likely scenarios in which the model performs mostly "look around" actions. The following 80\% of predictions lie within 20 meters, a maximum distance representative of the majority of the training data. There exists a taper around the 5 meter mark, followed by a rise around 10 meters and another taper after 15 meters. We believe this indicates the majority of self-control actions are of a reasonable distance magnitude. This is shown on Figure \ref{fig:cdf-of-spatial-action}.

\section{Additional Temporal Results}

\begin{figure}[!h]
\centering
    \includegraphics[width=0.75\linewidth]{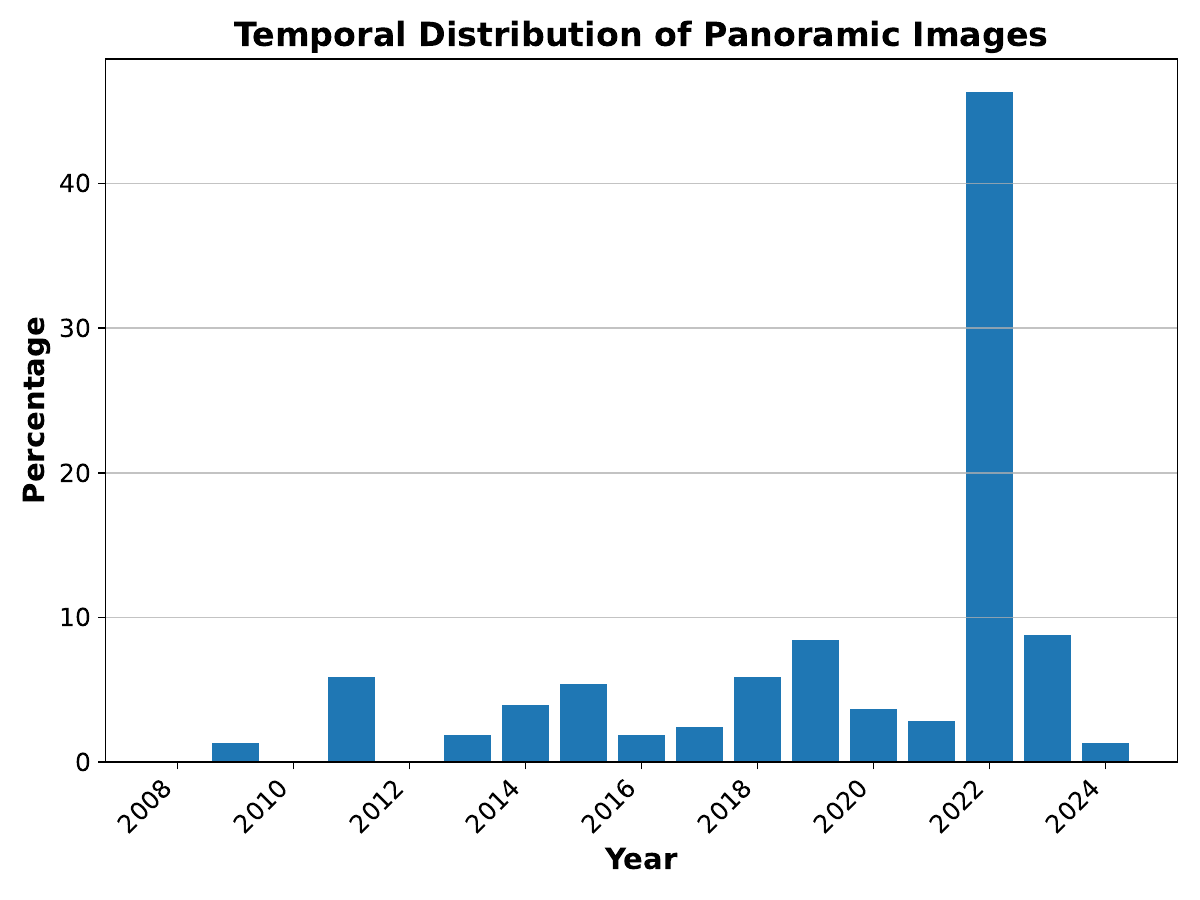}
    \caption{
    \textbf{GSV Data Temporal Distribution.} 2008 to 2022 form part of the training set, 2023-2024 are fully within the testing set. Although this data is unbalanced to a degree we do not observe large perplexity changes when evaluating on non 2022 data.}
    \label{fig:temporal-distribution-of-streetviews}
\end{figure}

\begin{figure}[H]
\centering
    \includegraphics[width=0.75\linewidth]{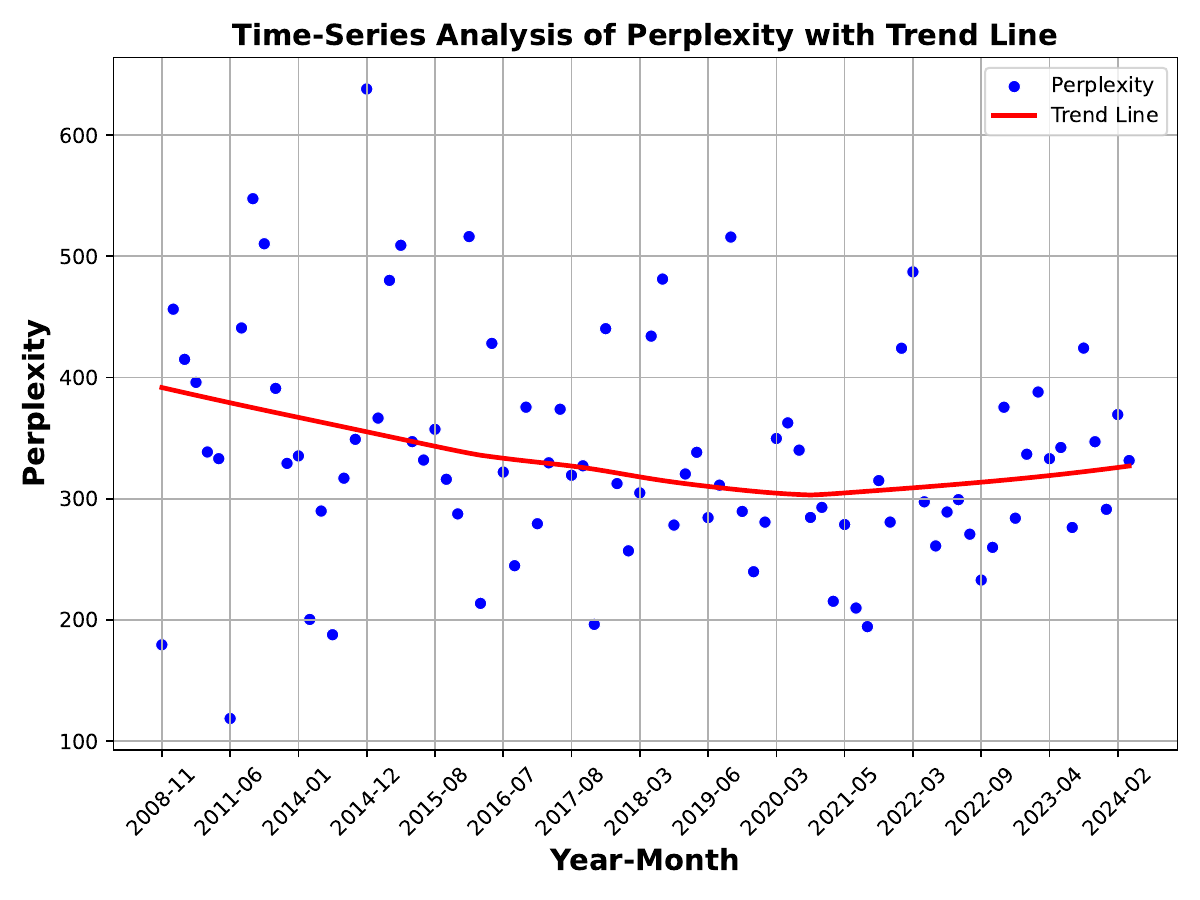}
    \caption{\textbf{Perplexity vs Year-Month Dates.} Across a range of dates average perplexity remains mostly consistent.}
    \label{fig:time-series-analysis-of-perplexity-with-trend-line}
\end{figure}

We measure generated image perplexity against time, including our held-out years (2023, 2024). We note this evaluation, as all others on our paper, is performed on the test set. We observe a decreasing perplexity trend as we reach the year 2022, which is overrepresented on the training data (shown on Figure \ref{fig:temporal-distribution-of-streetviews}). We do not observe a significant error trend increase when generating images into our fully held-out years. This is visualized on Figure \ref{fig:time-series-analysis-of-perplexity-with-trend-line}.

\begin{figure}[H]
\centering
    \includegraphics[width=0.75\linewidth]{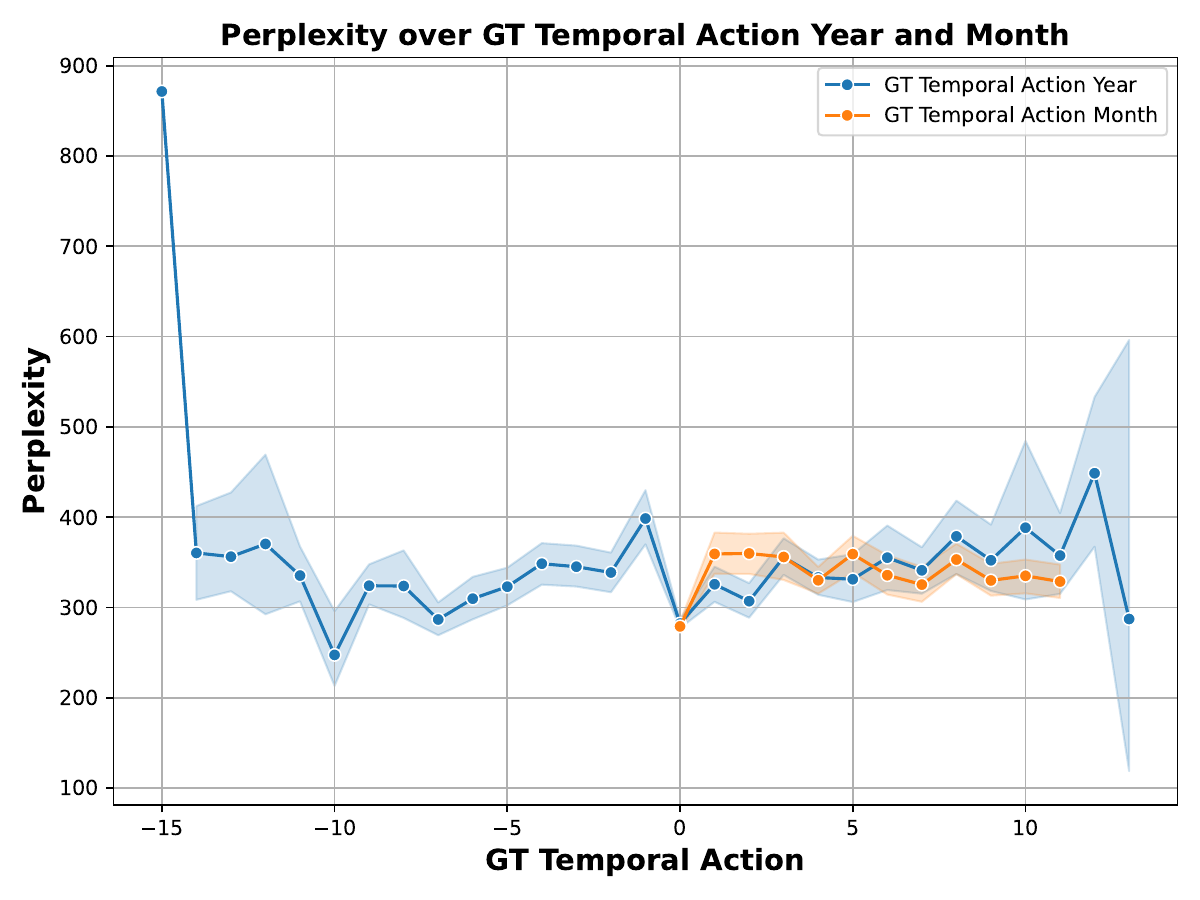}
    \caption{\textbf{Perplexity over GT Temporal Action Year and Month.} Non-extreme temporal instructions show a consistent perplexity score.}
    \label{fig:perplexity-over-gt-temporal-action-year-and-month}
\end{figure}

Additionally, we measure perplexity as a function of temporal actions. We observe the largest perplexity error when very large commands are given, for example -15 years. For the full range of month change commands, we observe perplexity mostly remains consistent. A visualization of this effect is found on Figure \ref{fig:perplexity-over-gt-temporal-action-year-and-month}.

\begin{figure}[!h]
\centering
    \includegraphics[width=0.75\linewidth]{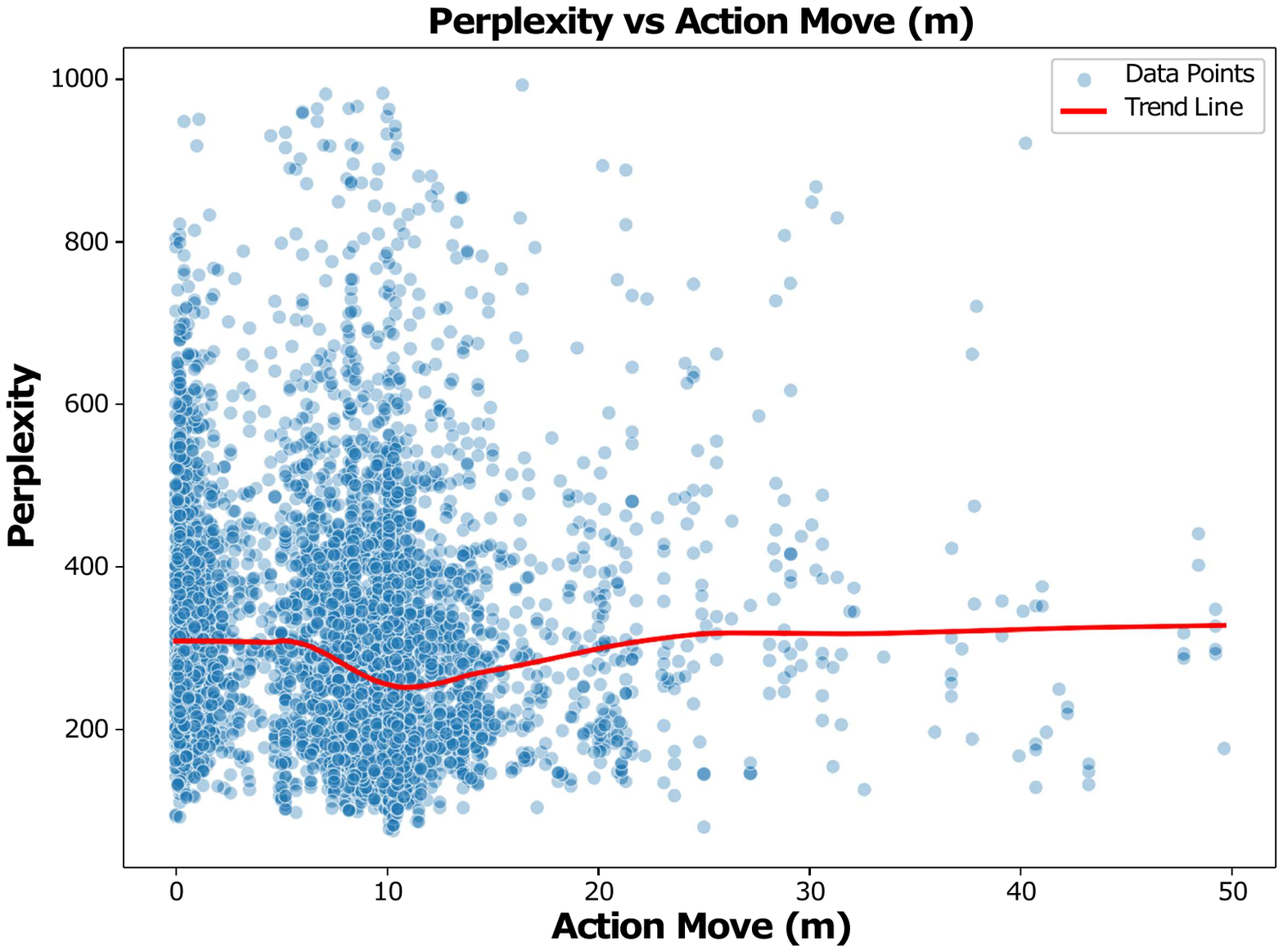}
    \caption{\textbf{Perplexity vs. Testing Action Move.} The majority of temporal instructions show a consistent perplexity score.}
    \label{fig:perplexity-over-Testing-action-move}
\end{figure}

Finally, we quantify perplexity under different commanded move action magnitudes. We find the trend line is mostly consistent, with the lowest error being found around the 10 meter action mark. This is likely due to the average distance between nodes in our training data is 7.5 meters. This can be found on Figure \ref{fig:perplexity-over-Testing-action-move}.

\section{Code and Data Release Documentation}
\label{app:code}


\subsection{Code Usage Instructions}
The code is thoroughly documented and released at \href{https://github.com/tera-ai/tardis}{https://github.com/tera-ai/tardis} under the Apache 2.0 License, with instructions for the installation, training, and testing of TARDIS. Example scripts for training are made available for both single-machine and distributed setups (using Kubernetes). An evaluation script is also included, along with a notebook with a step-by-step walkthrough of the whole evaluation process. We plan to release a more detailed code description post-publication.

\subsection{Dataset Format and Access}
The dataset can be downloaded and inspected (using the Dataset Viewer feature on HuggingFace) by following the instructions at \href{https://huggingface.co/datasets/Tera-AI/STRIDE}{https://huggingface.co/datasets/Tera-AI/STRIDE}. The dataset itself consists of two separate JSONL (line-delimited JSON) files, one for training and one for testing, which can be downloaded from publicly accessible Google Cloud Storage blobs through the use of the open-source Google Cloud CLI tool. Each JSON consists of a simple array of integers which themselves represent the tokens encoding a sample path or "visual sentence" of navigable spatiotemporal nodes. It is licensed under the Creative Commons Attribution Non Commercial 4.0 International Public License (CC-BY-NC-4.0). Only a portion of the complete dataset (10k out of 5M visual sentences) was detokenized in order to facilitate visualization through HuggingFace's Dataset Viewer, though we plan to upload the complete version post-publication for a more comprehensive experience.

\end{document}